\documentclass[journal=jcisd8,manuscript=article]{achemso}

\usepackage{chemformula} 
\usepackage[T1]{fontenc} 

\usepackage{algorithm}
\usepackage{algpseudocode}

\usepackage{graphicx}     
\usepackage{subcaption}   
\usepackage{adjustbox}    

\usepackage[table]{xcolor}
\usepackage[dvipsnames]{xcolor}

\usepackage{chngcntr} 
\usepackage{caption}  

\author{András Formanek}
\email{aformane@esat.kuleuven.be}
\affiliation[KUL]{Department of Electrical Engineering (ESAT), STADIUS Center for Dynamical Systems, Signal Processing and Data Analytics, KU Leuven, 3001 Leuven, Belgium}
\alsoaffiliation{Department of Artificial Intelligence and Systems Engineering, Faculty of Electrical Engineering and Informatics, Budapest University of Technology and Economics, Műegyetem rkp. 3, H-1111 Budapest, Hungary}

\author{Anna Vincze}
\affiliation{Department of Chemical and Environmental Process Engineering, Faculty of Chemical Technology and Biotechnology, Budapest University of Technology and Economics, Műegyetem rkp. 3, H-1111 Budapest, Hungary}
\alsoaffiliation{Center for Pharmacology and Drug Research \& Development, Semmelweis University, Üllői Str. 26, H-1085 Budapest, Hungary}
\alsoaffiliation{Department of Pharmaceutical Chemistry, Semmelweis University, Faculty of Pharmaceutical Sciences, Hőgyes Endre Street 7-9, H-1092 Budapest, Hungary}

\author{Richárd Bicsak}
\affiliation{Department of Chemical and Environmental Process Engineering, Faculty of Chemical Technology and Biotechnology, Budapest University of Technology and Economics, Műegyetem rkp. 3, H-1111 Budapest, Hungary}

\author{György T. Balogh}
\affiliation{Department of Chemical and Environmental Process Engineering, Faculty of Chemical Technology and Biotechnology, Budapest University of Technology and Economics, Műegyetem rkp. 3, H-1111 Budapest, Hungary}
\alsoaffiliation{Center for Pharmacology and Drug Research \& Development, Semmelweis University, Üllői Str. 26, H-1085 Budapest, Hungary}
\alsoaffiliation{Department of Pharmaceutical Chemistry, Semmelweis University, Faculty of Pharmaceutical Sciences, Hőgyes Endre Street 7-9, H-1092 Budapest, Hungary}

\author{Yves Moreau}
\author{Ádám Arany}
\affiliation[KUL]{Department of Electrical Engineering (ESAT), STADIUS Center for Dynamical Systems, Signal Processing and Data Analytics, KU Leuven, 3001 Leuven, Belgium}

\title[Comparative PAMPA QSPR Study]
  {A Comparative Study of QSPR Methods on a Unique Multitask PAMPA dataset 
  }

\abbreviations{QSPR, PAMPA, BBB}
\keywords{Quantitative Structure Property Relationship, Parallel Artificial Membrane Permeability Assay, Molecular Representation for Deep Learning}

\newcommand{\mytextsize}[2]{{\fontsize{#1}{0}\selectfont#2}}
\definecolor{revised_text}{rgb}{.0, .0, .0}
\definecolor{revised_bg}{rgb}{1., 1., 1.}

\begin{document}

\begin{tocentry}
\includegraphics[width=\textwidth, trim={1.2cm 1cm 1cm 1.2cm},clip]{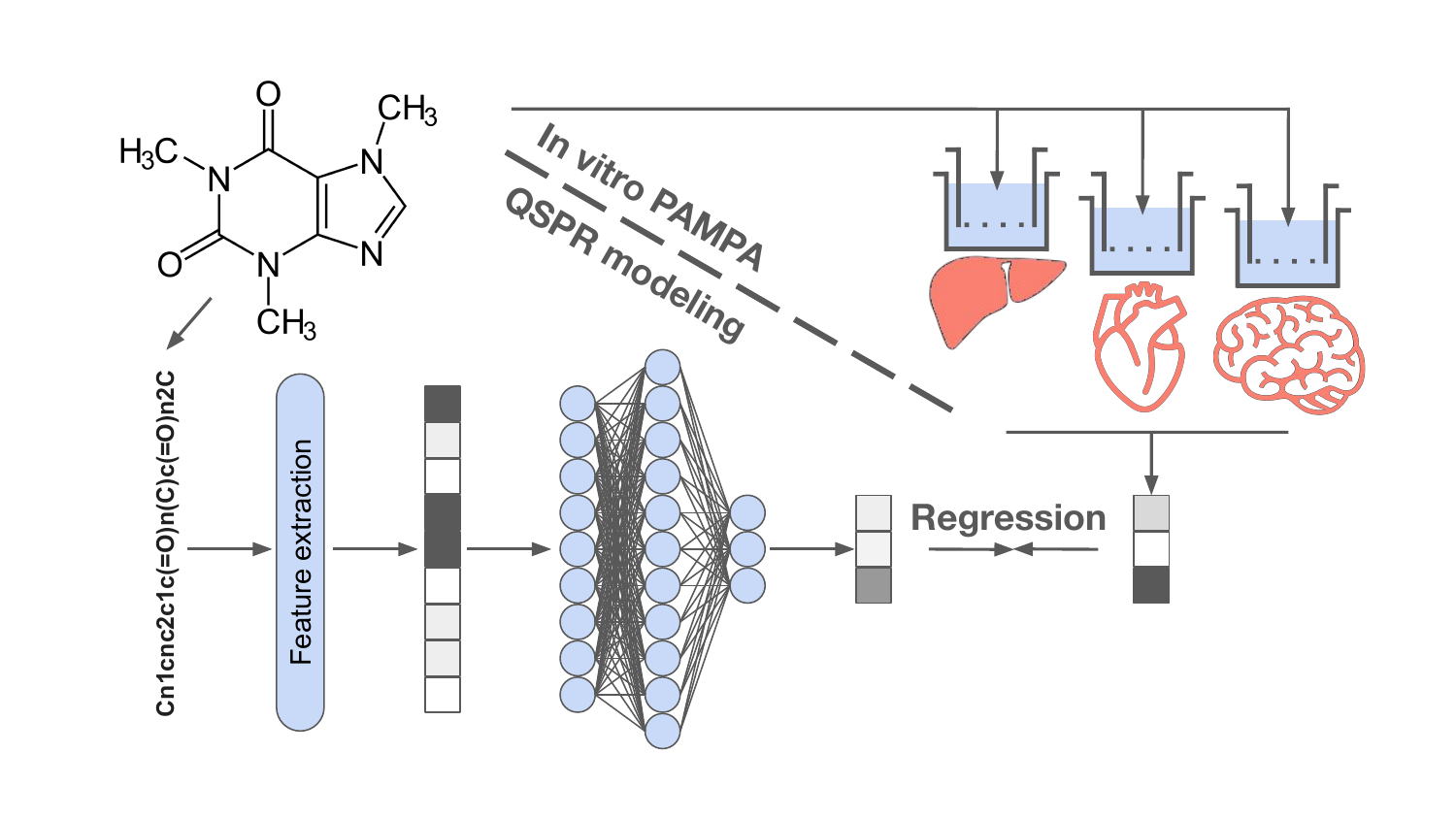}
\end{tocentry}

\newpage
\begin{abstract} 

We present a unique, multitask dataset comprising 143 drug and drug candidate molecules, 
each evaluated on in vitro, parallel artificial-membrane permeability assays (PAMPA) using six different model membranes.
Using this resource, we systematically assess the effectiveness of various molecular descriptors and regression models in predicting passive membrane permeability. 
The studied models range from simple linear regression to a modern pre-trained transformer architecture.
Particular attention is given to the trade-off between predictive performance and model interpretability, highlighting the challenges introduced by machine learning approaches. 
To our knowledge, this is the most comprehensive study on simultaneous modeling of multiple organ-specific PAMPA membranes to date, 
offering novel insights into membrane-specific permeability profiles.

We found that expert-designed physico-chemical property descriptors are more fitting for a limited sample size permeabilty study than deep learning based representations.

\end{abstract}

\section{Introduction}

Testing the ability of a drug to penetrate cell membranes is a key step of early drug discovery, as distribution to the pharmaceutically relevant organs, tissues, or cells is a prerequisite for therapeutic effect. 
Drug molecules may pass through biological membranes via active or passive transport, and the contribution of these mechanisms often varies according to the route of administration
\cite{sugano2010coexistence}. 
As a primary mechanism, passive transport is considered the most common route of intestinal absorption
\cite{tannergren2009toward} 
and blood-brain barrier penetration
\cite{di2012evidence} 
for lipophilic small molecules. 
Drug discovery relies on both computational models (\textit{in silico}) and experimental (\textit{in vitro}) assays to assess the membrane permeability of drug candidates. 
The \textit{in vitro} Parallel Artificial Membrane Permeability Assay (PAMPA)\cite{pampa1998}, 
which is best suited for high-throughput drug penetration screening,
was introduced three decades ago and has since been successfully incorporated into pharmaceutical research and development as an effective tool
\cite{jacobsen2023commercially}. 
In this technique, a porous polymer filter is sandwiched between two multi-well microtiter plates. 
Here, the artificial lipid membrane is fabricated by coating the polymer filter with a phospholipid (PL) solution, thus creating a cell membrane-mimicking surface between the two (donor and acceptor) aqueous compartments.
The major advantages of PAMPA are the high versatility in the possible membrane compositions combined with robustness and low cost. 
Since the first such setup using egg lecithin as artificial membrane 
\cite{pampa1998}, 
a number of variants were introduced 
ranging from simplified, PL-free/solvent-based assays (with solvents like 
hexadecane 
\cite{wohnsland2001high}, 
dodecane
\cite{enikHo2015vitro})
, others
using well-defined single PLs,
\cite{avdeef2001drug} 
or PLs combined with cholesterol 
\cite{sugano2001optimized}, 
all the way to natural PL mixtures extracted from animal tissues (brain polar lipid extract of porcine origin, modeling the blood brain barrier 
\cite{di2003high}) 

In 2006, Polli et al. investigated the effect of lipid composition on permeability across different PAMPA membranes
\cite{seo2006lipid}. 
They investigated five compounds on six glycerophospholipids with various acyl chains and head group charges. However, the narrow selection of compounds limited the generalizability of their conclusions. 
A previous study attempted to compare commercially available `total’ and `polar’ extracts of animal brain, heart and liver by testing 32 diverse drugs in PAMPA, and evaluate the dataset with the use of pharmacokinetic descriptors. 
\cite{vincze2023natural} 
To continue exploring the effects of phospholipids, our study includes 143 drug-like small molecules (including approved drugs) with 6 different PAMPA setups, featuring one solvent model (dodecane), two single lipids carrying different net charges (phosphatidylcholine - neutral, phosphatidylserine - negative), and three natural polar lipid extracts (brain, heart, and liver). 

Dodecane, a hydrocarbon, is the most used PL-solving agent in PAMPA assays, as it may closely mimic the bilayer’s hydrocarbon domain in membranes.  
\cite{mayer2002transport}. 
In some cases it even proved to be the most predictive model used as a single-component artificial membrane. 
\cite{enikHo2015vitro}. 
In a previous study Müller et al. investigated the effects of various phospholipid to dodecane ratios in BBB-PAMPA, showing that a lower dodecane ratio is needed to achieve a lipid-driven transport in a tissue-specific assay 
\cite{muller2015tuning}. 
Based on this evidence, dodecane acts as an individual barrier, therefore it has been chosen for our comparative study. 
To date, of all the published PAMPA setups, BBB-PAMPA has been the most applied and cited \textit{in vitro} assay
thanks to the commercial availability of Brain Polar Lipid Extract (BPLE) 
\cite{di2003high}. 
Based on the data available for total and polar lipid extracts, polar lipid extracts are richer in glycerophospholipids, which have been robustly used in PAMPA assays for decades.
\cite{vincze2023natural}
Therefore, along with brain polar extract, the polar extracts of heart and liver were also investigated in this study. 
Phosphatidylcholine (PC) was chosen as one of the single lipids investigated, as it is one of the main components of biological membranes, and the membrane component used for our corneal-specific PAMPA model 
\cite{dargo2019corneal}. 
To make our study more diverse, negatively charged phosphatidylserine (PS) was chosen as another single lipid species. 
As a significant component of brain polar lipid extract, PS presumably plays a key role in brain-specific permeability.

Quantitative Structure Property Relationship (QSPR) mathematically links the structure of molecules with physical or chemical properties.
This computational approach relies on the following three \cite{boczar2024review,hu2010review}
essential components: 
\begin{enumerate}
    \item \textit{In vitro} datasets, such as those obtained from PAMPA assays, which provide the experimental foundation for modeling.
    \item Molecular descriptors that capture the structural and/or physicochemical characteristics of compounds. 
    \item Modeling techniques, ranging from simple linear regression models to advanced machine learning techniques, such as neural networks, to establish predictive relationships. 
\end{enumerate}
The success of QSPR depends on the synergy between these components: new \textit{in vitro} data that extends the knowledge base, identification of the most informative descriptors, and modern ML techniques that can model complex, nonlinear relationships between descriptors and target properties. 

Since the goal of the modeling exercise is to later accurately predict the property of interest for new molecules, a realistic assessment of the true predictive power of the model must be established.\cite{hu2010review, fourches2010trust}
Rigorous validation, both internal (e.g., cross-validation within the training dataset) 
and external (using independent dataset), ensures the robustness and generalizability of these models. 
The integration of these elements enables more accurate predictions, supporting applications in 
drug discovery \cite{TIAN2020103888,oldenhof2023industry}
(e.g., predicting ADMET properties), material science \cite{ding2021molecular, quadri2022development}, and environmental chemistry \cite{thomas2019silico,nolte2017review}, while reducing reliance on resource-intensive trial-and-error methods. 
Encouraging collaborative efforts to publish new experimental datasets, refine molecular descriptors, and explore innovative algorithms can steadily improve the predictive capabilities of QSPR, reinforcing its role as a valuable tool for studying and predicting molecular properties.

\section{Materials and Methods}

\subsection{Materials}

Analytical grade solvents like acetonitrile, hexane, dodecane, and chloroform were purchased from Merck KGaA (Darmstadt, Germany). 
Most of the active pharmaceutical ingredients (APIs) were purchased from Merck KGaA (Darmstadt, Germany) and Mcule.com Ltd (Budapest, Hungary), while some of them (compounds PGY0072, PGY0216, HK275, HK363, HK416, HK814, HK990, VB253, MOE0448; see SMILES codes in the Supporting Information) were provided by Egis Pharmaceuticals Plc (Budapest, Hungary). 
Heart (H) and liver (L) polar lipid extracts of bovine origin, brain polar lipid extract (BBB), and L-$\alpha$-phosphatidylserine (PS) of porcine origin were purchased from Avanti Polar Lipids Inc. 
(Alabaster, AL, USA). 
\footnote{https://avantilipids.com/product-category/natural-lipids/extracts}
L-$\alpha$-phosphatidylcholine (PC) was purchased from Merck KGaA (Darmstadt, Germany). 
Phosphate buffered saline (0.01 M) was prepared from a premixed powder also purchased from Merck KGaA (Darmstadt, Germany). 
For all experiments, including HPLC analysis, water was provided by a Millipore-MilliQ water purification system.

\subsection{In vitro permeability assay (PAMPA)}

\textit{In vitro} PAMPA measurements were carried out as follows.
For each plate, 16 mg PL or PL extract were weighted and dissolved in 600 $\mu$L solvent mixture of chloroform, hexane, and dodecane (5:70:25 $v/v\%$) at $0^{\circ}$C, and filter membranes of the donor wells were coated with 5 $\mu$L lipid solution, or dodecane alone. 
The acceptor wells contained phosphate buffered saline (PBS, pH $7.4$; $0.01$M sodium phosphate, $0.138$M sodium chloride; $0.0027$M potassium chloride), and the APIs were dissolved in the same medium (donor wells). 
After 4 hours incubation at $35^{\circ}$C, all samples were analyzed by HPLC-DAD. 
The membrane retention (MR, Equation \ref{eq:membrane_retention} ) and effective permeability ($P_e$, Equation \ref{eq:permeability}) 
of the APIs were calculated as follows: \cite{avdeef2012permeabilityb}
\begin{align}
\label{eq:membrane_retention}
\text{MR} &= 1 - \frac{C_D(t)}{C_D(0)} - \frac{V_A c_A(t)}{V_D c_D(0)}\\
\label{eq:permeability}
\text{P}_e = \frac{-2.303}{A \cdot (t-\tau_{ss})} \cdot \frac{1}{1+r_v} \cdot \log&\left[-r_v + \frac{1+r_v}{1-\text{MR}} \cdot \frac{c_D(t)}{C_D(0)} \right] 
\end{align}
where $A$ is the filter area (0.3 cm$^2$), 
$V_D$ and $V_A$ are the volumes in the donor (0.15 cm$^3$) and acceptor phase (0.3 cm$^3$), 
$t$ is the incubation time (s), 
$\tau_{ss}$ is the time to reach steady-state (s), 
$c_D(t)$ is the concentration of the compound in the donor phase at time point $t$ (mol/cm$^3$),
$C_D(0)$ is the concentration of the compound in the donor phase at time point zero (mol/cm$^3$),
$c_A(t)$ is the concentration of the compound in the acceptor phase at time point $t$ (mol/cm$^3$),
$r_v$ is the aqueous compartment volume ratio ($V_D/V_A$).

\subsection{HPLC analysis}

The samples were analyzed using an Agilent $1100$ HPLC system equipped with a solvent mixer and quaternary pump, autosampler, column thermostat, and a DAD detector module (Agilent Technologies Inc., Santa Clara, CA, USA). 
As a stationary phase, a Kinetex C18 column (3 $\times$  30 mm, 2.6 $\mu$m) was used; 
the temperature was kept at $45^{\circ}$C. 
Two mobile phase solvents were used in a 3.6 minute-long gradient program: 
water, containing 0.1 $v/v\%$ formic acid (A) 
and AcN:water 95:5, containing 0.1 $v/v\%$ formic acid (B).
At the start of the gradient program, 
the column was flushed with eluent $2\%$ B for 0.3 min 
and then it reached $100\%$ B within 1.5 min, 
100\% B was kept for another 0.6 min, 
and then B dropped to $2\%$. 
Chromatograms were collected and processed with the ChemStation software (Version B.04.03.); 
peaks were integrated either at wavelength 220 nm or 254 nm based on spectral characteristics.


\subsection{Compound library}
The final data set consists of 143 compounds collected from 3 different sources.
One batch consists of 16 compounds selected to be handled together.
Every such compound was measured on 6 targets (described below) with 3 repeats on the same plate (2 targets per plate), resulting in 3 physical 96-well plates for each batch. 

The first 7 batch of measurements (112 molecules) were executed on 
\color{revised_text}
a diversity selected subset of 
\color{black}
the BUTE: Biomimetic Technologies Research Group's in-house dataset of drug-like molecules
\color{revised_text}
created by maximizing Tanimoto distance between ECFP representation of the selected compounds.
\color{black}
After the measurement results of the first 6 plates were at hand, we used experimental design techniques to select additional samples from the compound set provided by Mcule.com Ltd\cite{mcule}.
This is a large curated database containing commercially available compounds.
In our case, the list consisted of 52,591 possible molecules, 9,333 of which passed simple tests of drug-likeness 
and exhibited detectable chromophores, an essential feature for UV-based detection, in our analytical setup.

RDKit and Percepta \cite{landrum2013rdkit,percepta} descriptors (described below) were computed for these and the 21 most important descriptors were chosen by fitting linear models (linear regression, lasso, ridge, PLS) with forward feature selection on the single targets and the 1st PCA component.
Using only these chosen 21 descriptors, 32 of the 9,333 molecules were chosen to be purchased by D-optimal diversity selection. 
Because of the unavailability of some selected compounds and solubility issues, we replaced them with new ones, offered by SU: Department of
Pharmaceutical Chemistry research group, using the same D-optimal diversity selection algorithm as before.

\subsection{Dataset}

All molecules are initially represented by their SMILES \cite{smiles1988,smiles2017} (Simplified Molecular Input Line Entry System).
Although the SMILES of a molecule is not unique, RDKit offers a canonical version.
Removal of salts (low molecular  weight counter ions) and a general standardization of the molecules was carried out using RDKit version 2023.09.6 according to Algorithm \ref{alg:desalt} in Supporting Information.

\subsubsection{Derived descriptors}

Based on the desalted SMILES, five different numerical representations of the compounds were derived with various sizes and levels of interpretability.

\textbf{Percepta} 
by ACD/Labs \cite{percepta}
can be used for predicting various molecular properties and generating chemical descriptors for compounds.
It is widely used in cheminformatics and computational drug discovery
\cite{cornelissen2023BBB, shi2022IntestinalAbsorption,vincze2021corneal,ribeiro2017determination}
for its accuracy in predicting a range of physicochemical, ADME, and toxicity properties, such as logP, logD, logS, pK$_a$, Caco-2, which are highly relevant in the context of the present study\cite{tran2023recent}.
In our research, 57 such properties have been computed for all molecules.
After filtering and preprocessing, 38 of those descriptors were kept in the dataset.
A full list is available in the Supporting Information.
The precomputed Percepta descriptors are also available in the Supporting Information file `percepta.csv'.

\textbf{RDKit} is a popular open-source cheminformatics library \cite{landrum2013rdkit} also widely used in computational chemistry and drug discovery. 
\cite{cornelissen2023BBB,oldenhof2020chemgrapher,orosz2022comparison,tan2024predicting}
It provides tools for performing cheminformatics tasks, such as calculating molecular descriptors, such as molecular weight, topological polar surface area (TPSA), number of rotatable bonds, and basic physicochemical properties, such as logP. 
These descriptors are easy to interpret.
In our research, 96 such properties have been computed for all molecules.
A full list of these properties is available in the Supporting Information.
The precomputed RDKit descriptors are also available in the Supporting Information file `rdkit.csv'.

\textbf{ECFP} (Extended-Connectivity Fingerprints\cite{ecfp}), also known as circular fingerprints or Morgan fingerprints, is a widely-used molecular representation 
\cite{oldenhof2023industry,gao2021revolutionizing,di2023systematic,sawada2014benchmarking}
that encodes chemical structures into fixed-length binary vectors, suitable for machine learning models. 
ECFP6 focuses on local molecular environments, with a radius of 3 and is invariant to rotation and translation.
The ECFP6 input space is high-dimensional but vastly sparse.
In our paper, we folded \cite{le2020neuraldecipher} the fingerprint to 2,000.
The precomputed ECFP descriptors are also available in the Supporting Information file `ecfp.csv'. 
The lists contain the column indices of the ones in the raw vectors.

\textbf{CDDD} (Continuous and Data-Driven molecular Descriptors\cite{cddd})
was developed to overcome some limitations of traditional fingerprints, such as ECFP.
The descriptor vectors are generated as the inner representation of a sequence-to-sequence autoencoder neural network that is pretrained on a large chemical dataset in an unsupervised manner. 
This method generates continuous, fixed-length (512 dimensional) vectors, which encode molecular information in a way that is expected to be more suitable than discrete fingerprints for machine learning tasks involving search by similarity or clustering.
CDDD captures complex molecular relationships for deep learning models to better capture and predict compound activity behaviors. The precomputed CDDD descriptors are also available in the Supporting Information file `cddd.csv'.

\textbf{MolBERT} \cite{molBERT2020}
is also a fixed length (768 in this case) latent representation of an autoencoder neural network model trained on vast amounts of molecular SMILES data.
Unlike CDDD, it uses a BERT-like transformer\cite{devlin2019bert} architecture  to learn contextualized representations of molecules by predicting masked atoms or substructures within SMILES strings.
MolBERT aims at bringing the advantages of Natural Language Processing models into the molecular domain, helping to advance virtual screening, lead optimization, and other tasks in computational drug discovery.
The precomputed MolBERT descriptors are also available in the Supporting Information file `MolBERT.csv'.

\subsubsection{Measurement matrix}

The Supporting Information File 'all\_plate\_measurements.csv' contains the raw data of all $143$ compounds, repeated $3$ times (resulting in $429$ measurement rows). 
The column names correspond to the molecules, the plate number of the experiment, and 18 columns of measurement values 
\{MR, P$_e$, logP$_e$ \} $\times$ \{BBB, Liver, Heart, Dodecane, Phosphatidylserine, Phosphatidylcholine\}.
The file 'all\_plate\_desalted\_smiles\_measurements\_folds.csv' only contains the average values of measurements and in an additional column the standardized and desalted SMILES descriptors of the compounds. 
In our study, the logP$_e$ values are of primary interest, so that all the regression models and PCA analyses are applied to the corresponding 6-column submatrix only.

\subsection{Modeling}

We selected a wide range of machine learning (ML) regression models for our analysis and executed an exhaustive hyperparameter search.
We used the Scikit-learn \cite{scikit-learn} (version 1.2.2) Python library to train classical ML models on the data set.
The modeling classes were as follows (for hyperparameters see Supporting Information):
 
\begin{itemize}
    \item \textbf{DecisionTreeRegressor}: decision tree regressor (DTR) 
    \item \textbf{RandomForestRegressor}: random forest regressor (RFR) ensemble of 1,000 DTR models 
    \item \textbf{ElasticNet}: linear regression implementation with $L_1$ (Lasso) and $L_2$ (Ridge) type regularization.
    \item \textbf{MultiTaskElasticNet}: joint optimization of the ElasticNet to multiple targets 
    \item \textbf{BayesianRidge}: fits a Bayesian Linear Regression with Ridge ($L_2$ type) regularization
    \item \textbf{PLSRegression}: Partial least-squares regression \cite{PLS1986} commonly used \cite{vincze2021corneal} cross-decomposition algorithm in chemoinformatics
    \item \textbf{SVR}: Epsilon-Support Vector Regression model
    \color{revised_text}
    \item \textbf{XGBoost}: parallel (regression) tree gradient boosting model \cite{chen2016xgboost,xgboost-python}
    \color{black}
    
    \item \textbf{MLP}: finally, the SparseChem\cite{sparsechem2022} model is a Multi-Layer Perceptron (\textbf{MLP}) implementation specialized for sparse input and output matrices, developed, among others, for ML in chemistry tasks, such as Drug-Target Interaction (DTI) prediction. 
\end{itemize}    

From an interpretability point of view, MLP models are considered black boxes, but their prediction superiority (as we also show later) gives them a leading role in modern chemistry and drug discovery research.

\section{Experiments and results}

\subsection{Principal Component Analysis of membrane profiles}

\begin{figure}[ht]
    \centering
    \includegraphics[width=0.49\textwidth, trim={1cm 5cm 1cm 5cm},clip]{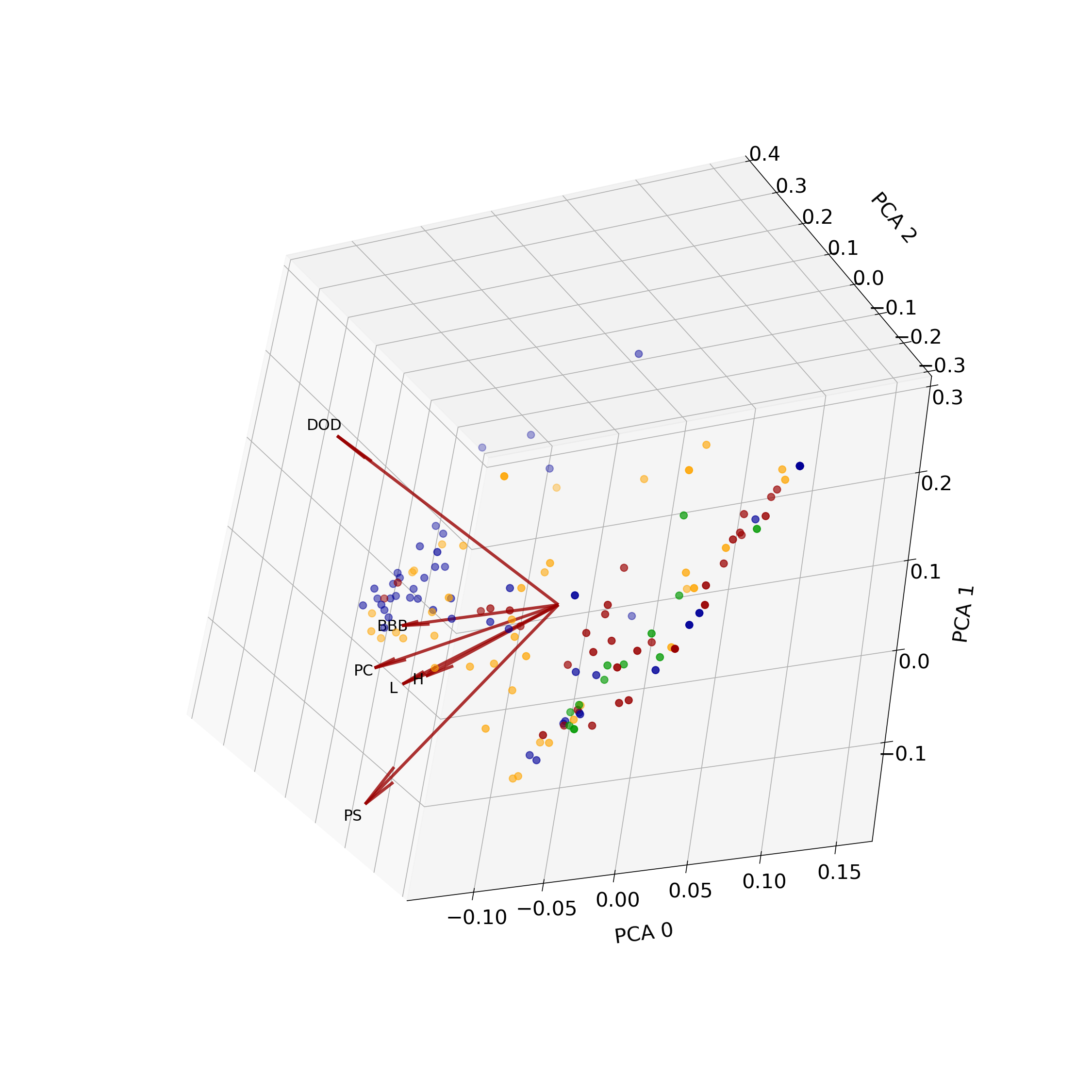}
    \includegraphics[width=0.49\textwidth]{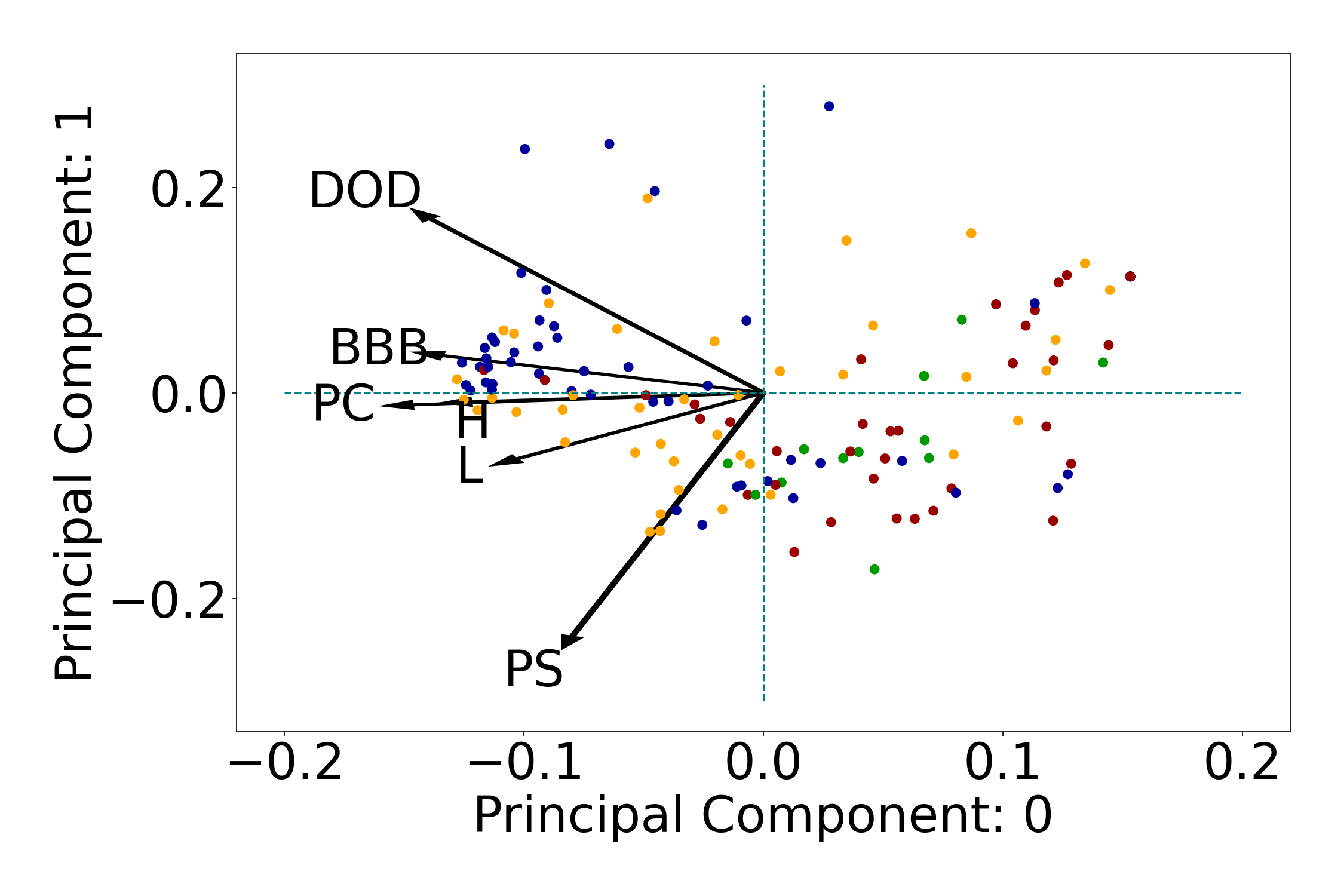}
    \includegraphics[width=0.49\textwidth]{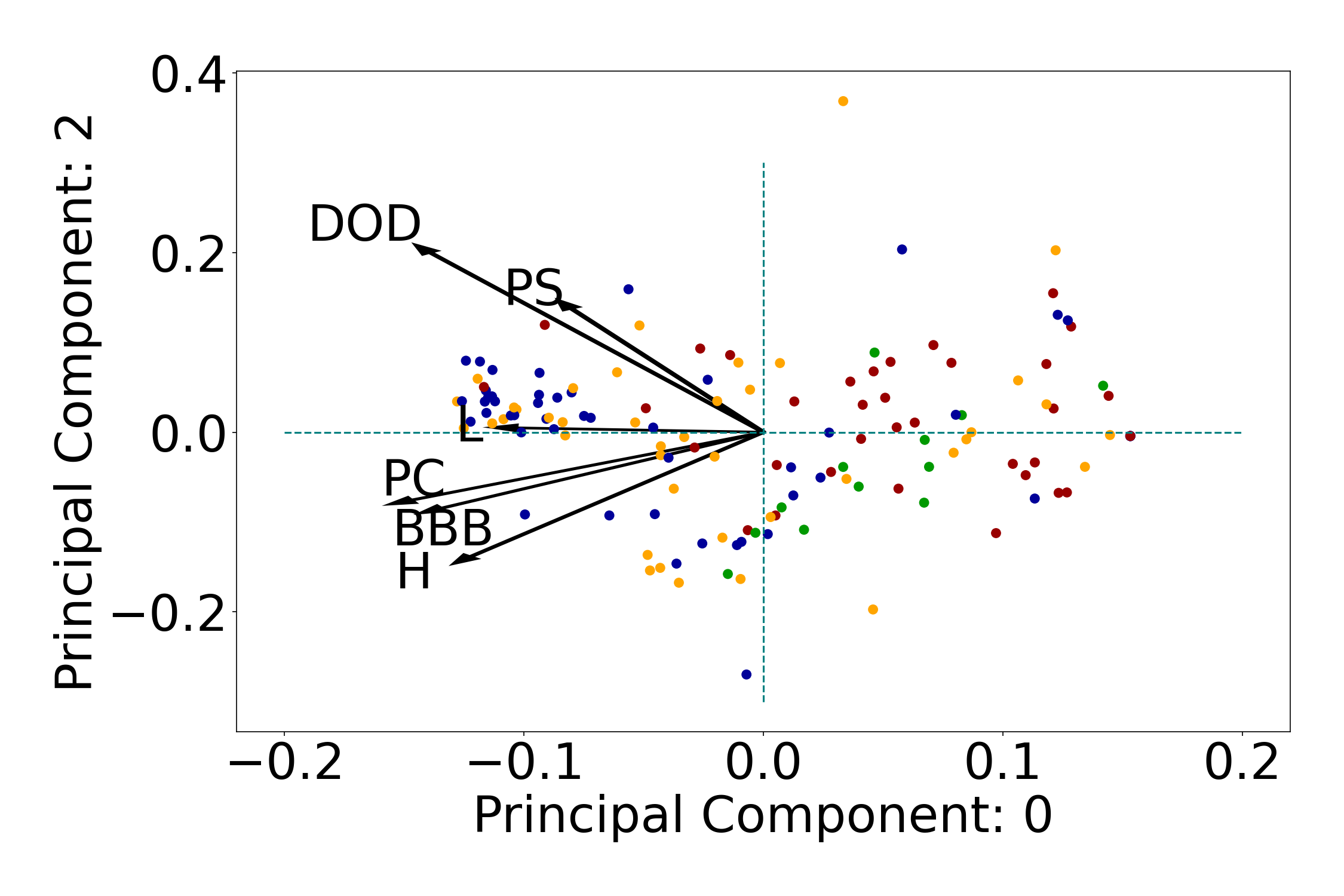}
    \includegraphics[width=0.49\textwidth]{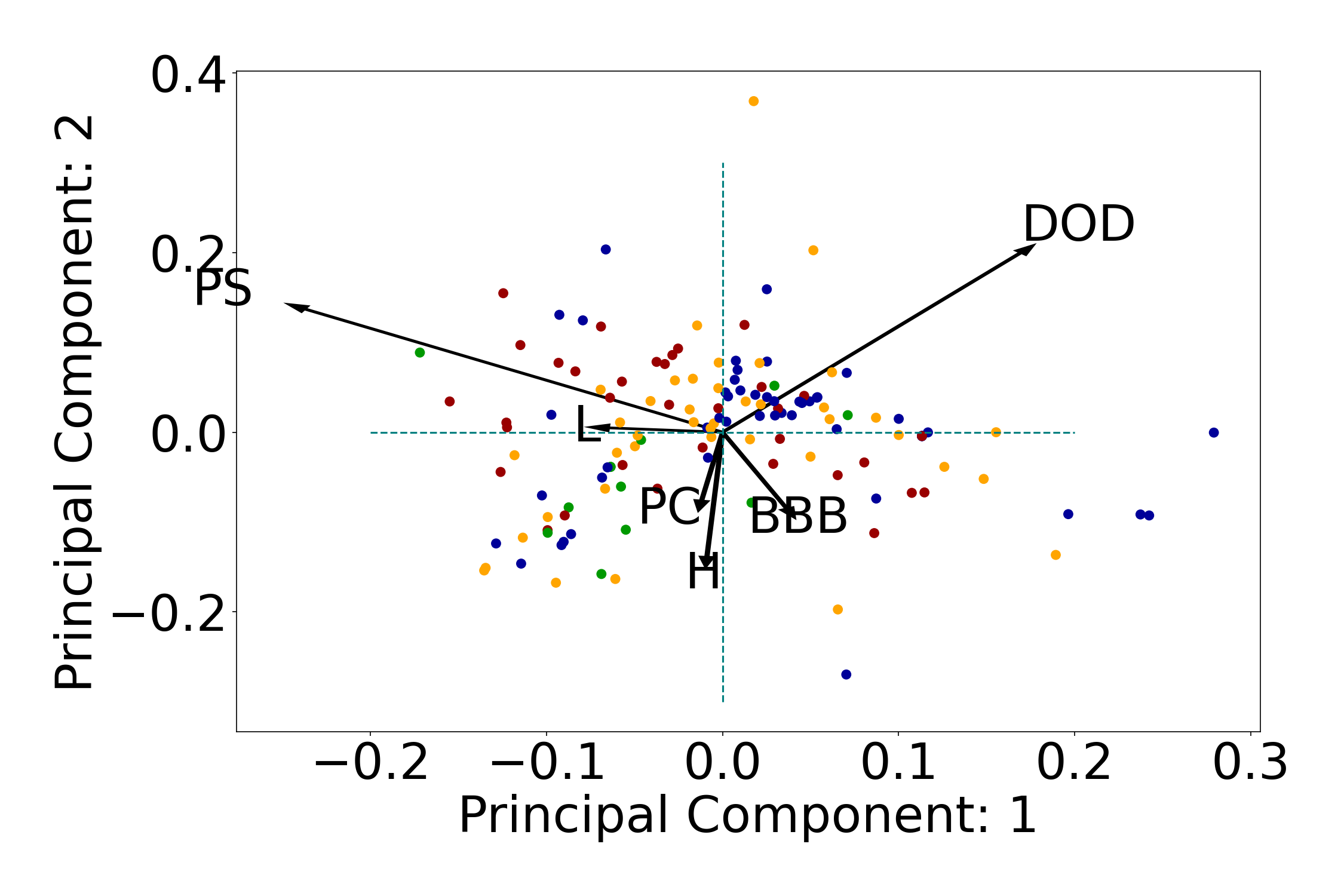}
    \caption{ 
    3D and 2D plots of the first 3 principal components of the matrix containing the measured logP$_e$ values for all molecules on all 6 membranes. 
    Colors are based on charge: 
    red  -  anionic, blue  -  cationic, green  -  zwitter ionic, yellow  -  neutral.
    }
    \label{fig:PCA01}
\end{figure}

We applied Principal Component Analysis (PCA) to the matrix containing the measured logP$_e$ values for all molecules on the six membranes (BBB, L, H, DOD, PS, PC).
The results of the PCA analysis are shown in Figure~\ref{fig:PCA01}.
Each dot represents a compound, colored according to their charge state at pH 7.4. 
The red color represents anionic (mainly acids), blue cationic (mainly bases), green zwitter ionic (mainly amphoteric compounds), and yellow neutral compounds (mainly compounds without ionizable moieties).
The six arrows correspond to the membranes, and they show the main directions where molecules, that cross the given membrane, lie.

We can see that the first principal component (PCA$_0$) is a good indicator of the permeability in general. 
Although, there is no perfect trend, the cations (bases) and neutral compounds are mainly located in the same domain where the
loading vectors point at, meaning that they have generally better permeability.
Meanwhile anions (acids) and zwitterions (amphoteric compounds) primarily populate the other end of the $PCA_0$ domain, as they are generally less likely to penetrate biological membranes via passive transport.
This can be explained by the fact that biological membranes are net neutral or partially negatively charged at physiological pH.
The selectivity effects of various membranes appear in the further principal components.
The first three principal components explain $65.88\%$, $10.67\%$ and $7.98\%$ of the variance respectively.
This indicates that, although there is a strong common factor in the permeability of different membranes, there is also value in modeling the membrane differences. 

It has to be mentioned that during the experiments with PS we experienced recurring inconsistencies in solution preparation, which resulted in a poorly predictable dataset. 
Based on this experience, PS may not have been the best choice for this research.
Future investigations should consider using different negatively charged PLs such as phosphatidylinositol or cardiolipin.

\subsection{QSPR modelling}

We performed exhaustive QSPR analysis for PAMPA membrane permeability (logP$_e$) across the different membranes using various molecule representations and regression models.
We split the dataset \textcolor{revised_text}{randomly into} $5$ equal size folds,
the first of which is held out for external testing purposes.
\color{revised_text}
Our scaffold analysis shows that the final compound set consists of 139 unique scaffolds.
(For overlapping scaffolds see Supplementary Table~\ref{tbl:scaffold_repeats}.) 
Therefore, random splitting is sufficiently conservative.
\color{black}

The regression models were all tuned using \textbf{4-fold cross-validation} (CV) using 3 folds as train and 1 fold as validation.
The train and validation metrics are averaged over the 4 cross-validation folds.
In addition, the MLP trainings were repeated 5 times within each CV setup, with different random initial weights, using early stopping with a maximum of 200 epochs.
The random forest trainings were repeated 3 times.
For other machine learning methods, the results are deterministic, therefore repeats were not meaningful.
In all cases, we optimized for Root Mean Squared Error (RMSE) during training and within each model category we chose the optimal set of hyperparameters with the highest coefficient of determination on the validation set ($R^2$ valid).
In case of multitask learning, the training was done on the average score of the 6 tasks, but model selection was done based on individual task validation performance.

\begin{figure}[t]
  \caption{
  Hierarchical scatter plot by input representation and target.
  Best validation R$^2$ of each model class and input-target pair.
  Hyperparameter search was done in 4-fold cross-validation. 
  For MLPs, the average of 5 repeats with random initialization is considered.}
  \label{fig:all_model_best}
  \fcolorbox{revised_bg}{revised_bg}{\includegraphics[width=0.95\linewidth]{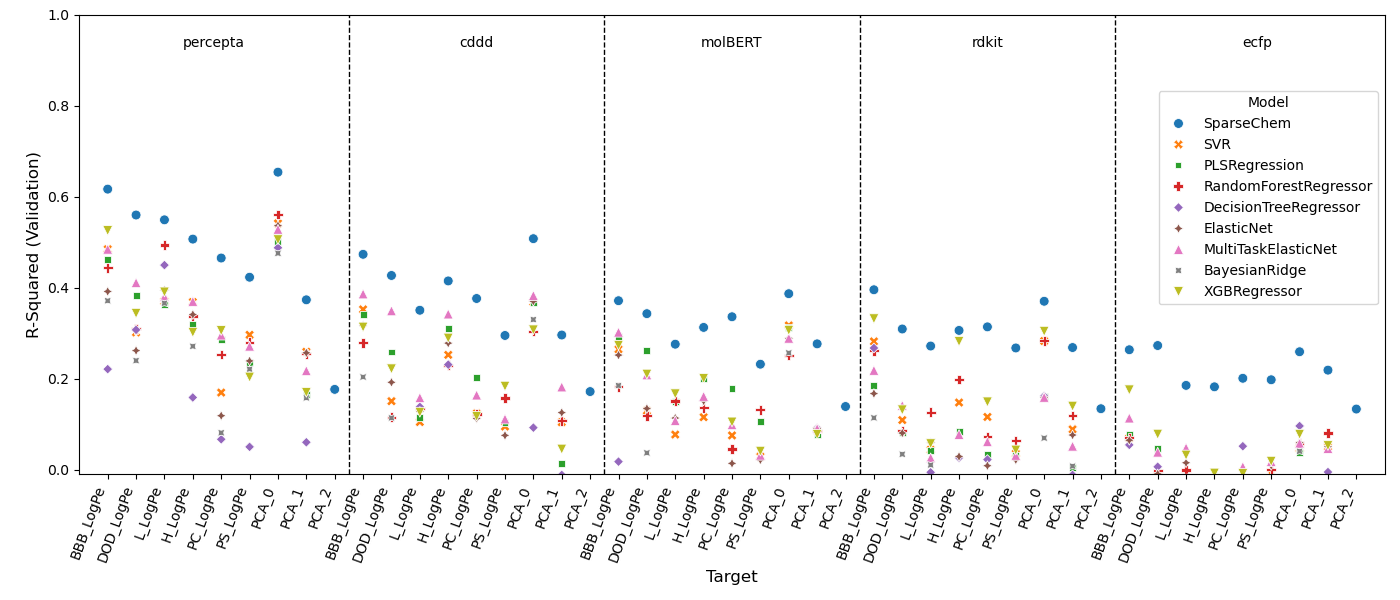}}
\end{figure}

We call the models with the best validation R$^2$  (after hyperparameter search) the \textit{tuned models} (for each model class and input-target pair).
In Figure~\ref{fig:all_model_best}, we visualize the validation R$^2$ of these models by model class (in different colors and shapes) grouped by input-target pairs (in columns).

As we can see the MLP models (blue dots) dominate all other model classes. Furthermore, it is remarkable that the usefulness of  feature representations is independent of the regression model class.
The feature representations in decreasing order of their descriptive power are Percepta, CDDD, MolBERT, RDKit, and ECFP.
In addition, independently of the model class, the best predictable target is PCA$_0$, which was previously established to be a common factor of the permeability of the various membranes.

As the MLP model trained on Percepta features dominates all other options for all target membranes, first, we present these results in Table~\ref{tbl:absolute_best},
and later show results of other noteworthy setups, emphasizing different aspects of the QSPR with respect to other model classes and input representations. 

In Table~\ref{tbl:absolute_best}, we show the \textit{tuned models} with the overall best validation performance for each target.
In all cases this was an MLP model trained on Percepta features in a multitask setting (6 targets).
For corresponding hyperparameters, see \textcolor{revised_text}{Table~\ref{sup:HP_absolute_best}} in the Supporting Information.
To avoid spurious conclusions, because of the large number of examined regression models,
we only evaluate on the external test set a limited set of \textit{tuned models}.
First, those models that have validation R$^2$ higher than 0.5 in Table~\ref{tbl:absolute_best} (bold). 
The results of which are presented later in Table~\ref{tbl:external_test}.

\begin{table}[]
  \caption{Overall best \textit{tuned models} on the validation set for each target after hyperparameter search with 4 folds cross-validation.
  The results of 5 repeats are used to calculate the metrics.
  For hyperparameters of the trained models see Table~\ref{sup:HP_absolute_best} in the Supporting Information.}
  \label{tbl:absolute_best}
  \begin{tabular}{lrrrr}
    \hline
    target & corr train & corr valid & $R^2$ train & $R^2$ valid \\
    \hline
    $\mathbf{BBB}$  & 0.8062 (0.0172) & 0.7979 (0.0073) & 0.6343 (0.0300) & \textbf{0.6169} (0.0106) \\
    $\mathbf{DOD}$  & 0.6717 (0.0133) & 0.7636 (0.0169) & 0.3883 (0.0462) & \textbf{0.5599} (0.0197) \\
    $\mathbf{L}$    & 0.7282 (0.0111) & 0.7526 (0.0252) & 0.4918 (0.0065) & \textbf{0.5493} (0.0308) \\
    $\mathbf{H}$    & 0.7267 (0.0138) & 0.7167 (0.0211) & 0.4964 (0.0204) & \textbf{0.5069} (0.0215) \\
    $PC$   & 0.6497 (0.0396) & 0.6912 (0.0169) & 0.3905 (0.0513) & 0.4653 (0.0240) \\
    $PS$   & 0.6401 (0.0488) & 0.6653 (0.0184) & 0.3659 (0.0874) & 0.4231 (0.0214) \\
    \hline
    $\mathbf{PCA_0}$ & 0.7862 (0.0230) & 0.8165 (0.0183) & 0.5946 (0.0405) & \textbf{0.6542 }(0.0243) \\
    $PCA_1$ & 0.6765 (0.0501) & 0.6205 (0.0179) & 0.4406 (0.0662) & 0.3733 (0.0192) \\
    $PCA_2$ & 0.4968 (0.0282) & 0.5109 (0.0327) & 0.2207 (0.0254) & 0.1763 (0.0202) \\
    \hline
\end{tabular}
\end{table}

Table~\ref{tbl:sklearn_best} shows the best non-neural network models for each target.
The Percepta representation is still the most informative in all cases.
Contrary to the clear dominance of the MLP in the previous case, there is no clear ordering of model classes here, although, less powerful models are generally dominated by their more flexible variants.
Furthermore, it is noteworthy that in Table~\ref{tbl:absolute_best} train and validation values are consistently closer to each other than in Table~\ref{tbl:sklearn_best}. 
This may seem counterintuitive at first glance. 
According to conventional wisdom, neural network models are more complex therefore should suffer more from overfitting. 
However, in the light of Supporting Information Table~\ref{sup:HP_absolute_best} the result is understandable. 
Dropout combined with weight decay serves as a powerful regularization, reducing effective model complexity so that the MLP models can avoid overfitting.
In Table~\ref{tbl:sklearn_best} only the validation R$^2$ of the Random Forest Regressor model for PCA$_0$ \color{revised_text} and the XGBoost model for BBB \color{black} is over 0.5, and therefore chosen for testing.

\begin{table}
  \caption{Best classical ML models for each target after hyperparameter search with 4 folds cross-validation. 
  For hyperparameters of the trained models see Table~\ref{sup:HP_sklearn_best} in the Supporting Information.
  }
  \label{tbl:sklearn_best}
    
    \begin{tabular}{llrrrr}
    \hline
    target & model & corr train & corr valid & $r^2$ train & $r^2$ valid \\
    \hline
    $L$    & RFR     & 0.8636 & 0.7291 & 0.7317 & 0.4969 \\
    $PS$   & SVR   & 0.8783 & 0.5833 & 0.7360 & 0.2964 \\
    $DOD$  & MTEN   & 0.7647 & 0.6845 & 0.5674 & 0.4110 \\
    \rowcolor{revised_bg}{$PC$}   & XGB   & 0.8364 & 0.5602 & 0.6526 & 0.3057 \\
    $H$    & MTEN   & 0.7609 & 0.6474 & 0.5707 & 0.3698 \\
    \rowcolor{revised_bg}{$BBB$}  & \textbf{XGB}   & 1.0000 & 0.7417 & 1.0000 & \textbf{0.5258} \\
    $PCA_0$ & \textbf{RFR} & 0.9699 & 0.7646 & 0.9159 & \textbf{0.5633} \\
    $PCA_1$ & SVR & 0.8945 & 0.5427 & 0.7460 & 0.2591 \\
    $PCA_2$ & SVR & 0.5115 & 0.3208 & 0.2467 & -0.0553 \\
    \hline
    
    \end{tabular}
\end{table}

To discuss the usefulness of input feature representations, we show in  Table~\ref{tbl:PCA_BBB_features} the best MLP models for each representation. BBB and PCA$_0$ are chosen as targets, because generally, these two are the best learnable tasks.
\color{revised_text}
Predicting PCA components serves the purpose to demonstrate how well universal features of membrane permeability versus organ selectivity can be predicted. Our focus is to examine the possibility for a tissue specific PAMPA model. On the other hand the majority of the existing literature do not contain studies of tissue specificity, therefore using the PCA$_0$ component allows us to link closer to existing literature while still emphasizing our focus on selectivity.
\color{black}

Finally, in Table~\ref{tbl:sklearn_PCA} the best non-neural network PCA$_0$ models are shown for each input feature representation.
This data also supports the observation based on Figure~\ref{fig:all_model_best}
about the general order of predictive power of the input representations.

We can ascertain that in our setting ECFP, which is a widely accepted gold standard in QSAR modeling, consistently underperformed the other input representations.
This can be explained with the high dimensionality of ECFP compared to the number of molecules in our training set, as the number of training samples is an order of magnitude smaller than the number of features (the dimensionality of the ECFP representation).

While the pretrained neural network model CDDD performs well, it cannot reach its full potential as a chemistry foundational model.  It is clear that one cannot avoid making a sufficient number of experiments even if one relies on a pretrained model. 
CDDD outperforms molBERT, a transformer-based foundational model.
Similarly as above, our observation is that sufficient training sample sizes is still needed. We do not see evidence for particularly good few-shot performance.

Note that MolBERT is on par with RDKit, which only contains explainable physicochemical and structural parameters.
We hypothesize that the main reason of the dominance of Percepta over RDKit is the lack of protonation models in the latter (i.e., the inability to predict $pK_{a}$ values, and any downstream pH dependent physicochemical parameters like logD).

The \textit{tuned models} with validation $R^2>0.5$ in Table~\ref{tbl:PCA_BBB_features} were already chosen previously, but to gain more insight on the other input representations, we selected further models from this Table by relaxing the threshold requirement.

\begin{table}[t]
    \caption{Predicting 1st PCA component of six membranes and the Blood-Brain Barrier penetration using different input representations. 
    Best MLP \textit{tuned models} are shown. 
    The results of 5 repeats are used to calculate the metrics.
    For hyperparameters of the trained models see Table~\ref{sup:HP_PCA_BBB_features} in the Supporting Information.
    }
    \label{tbl:PCA_BBB_features}
      \begin{tabular}{llrrrr}
        \hline
        representation & target & corr train & corr valid & $r^2$ train & $r^2$ valid \\
        \hline
        \textbf{percepta} & $PCA_0$ & 0.7862 (0.0230) & 0.8165 (0.0183) & 0.5946 (0.0405) & \textbf{0.6542} (0.0243) \\
        \textbf{cddd}     & $PCA_0$ & 0.8501 (0.0153) & 0.7300 (0.0134) & 0.6914 (0.0310) & \textbf{0.5079} (0.0185) \\
        \textbf{molBERT}  & $PCA_0$ & 0.6990 (0.0129) & 0.6455 (0.0223) & 0.4324 (0.0221) & \textbf{0.3868} (0.0155) \\
        rdkit    & $PCA_0$ & 0.5586 (0.0462) & 0.6242 (0.0138) & 0.2977 (0.0473) & 0.3702 (0.0155) \\
        ecfp     & $PCA_0$ & 0.7436 (0.0661) & 0.5422 (0.0249) & 0.4636 (0.0935) & 0.2591 (0.0180) \\
        \hline
        \textbf{percepta} & $BBB$ & 0.8062 (0.0172) & 0.7979 (0.0073) & 0.6343 (0.0300) & \textbf{0.6169} (0.0106) \\
        \textbf{cddd}     & $BBB$ & 0.7205 (0.0802) & 0.7043 (0.0169) & 0.5058 (0.1007) & \textbf{0.4735} (0.0208) \\
        \textbf{rdkit}    & $BBB$ & 0.6090 (0.0400) & 0.6512 (0.0153) & 0.3572 (0.0579) & \textbf{0.3955} (0.0210) \\
        molBERT  & $BBB$ & 0.5548 (0.0372) & 0.6291 (0.0165) & 0.2773 (0.0461) & 0.3714 (0.0085) \\
        ecfp     & $BBB$ & 0.6951 (0.0404) & 0.5454 (0.0256) & 0.4170 (0.0444) & 0.2635 (0.0288) \\
        \hline
    \end{tabular}

\end{table}

\begin{table}
  \caption{Predicting 1st PCA component using different input representations. 
  Best non-neural network \textit{tuned models} are shown.
  For hyperparameters of the trained models see Table~\ref{sup:HP_sklearn_PCA} in the Supporting Information.
  }
  \label{tbl:sklearn_PCA}
    \begin{tabular}{llrrrr}
    \hline
    input representation & model & corr train & corr valid & $r^2$ train & $r^2$ valid
 \\
    \hline
    percepta & RFR & 0.9699 & 0.7646 & 0.9159 & 0.5633 \\
    cddd & MTEN & 0.9488 & 0.6455 & 0.8558 & 0.3828 \\
    molBERT & SVR & 0.9559 & 0.6119 & 0.9098 & 0.3170 \\
    \rowcolor{revised_bg}
    rdkit & XGB & 0.9999 & 0.5905 & 0.9970 & 0.3043 \\
    ecfp & DTR & 0.5737 & 0.3965 & 0.3322 & 0.0958 \\
    \hline
    \end{tabular}
\end{table}

\subsubsection{Results on the external test compound set}

To ensure sufficiently high statistical power,  we chose to evaluate only a limited number of cases on the external test set
based on the validation R$^2$ values of the \textit{tuned models} presented above (Tables~\ref{tbl:absolute_best},~\ref{tbl:sklearn_best},~\ref{tbl:PCA_BBB_features},~\ref{tbl:sklearn_PCA}).
As was already outlined in the previous section, we first chose the overall best performing models, followed by the best non-MLP models all with validation R$^2 > 0.5$.
At this point all chosen models were still trained on Perceptra.
To learn about the other input representations, we added the best performing models based on Table~\ref{tbl:PCA_BBB_features} by relaxing the threshold requirement.
For visual explanation of the selections see Figure~\ref{fig:chose_model_visual} in the Supporting Information.

All but \color{revised_text}two \color{black} of the models selected according to these criteria were multitask MLPs. 
These were evaluated on the test compounds, which were excluded from the entire tuning procedure described above.
In Table~\ref{tbl:external_test}, we show the R$^2$ values of the evaluation on the external test compound set. For the corresponding correlation values, see the Supporting Information. 

\begin{table}
  \caption{
  Performance of the selected models, that correspond to various research questions, evaluated on the test set. 
  Ordered according to the validation $R^2$ values.
  }
  \label{tbl:external_test}
\begin{tabular}{lllrrr}
    \hline
    input    & target  & model & $R^2$ train & $R^2$ valid & $R^2$ test\\
    \hline
    
    percepta & PCA$_0$ & MLP & 0.5769 (0.0415) & 0.6435 (0.0142) & 0.4842 (0.0243)\\
    percepta & \underline{BBB}     & MLP & 0.6342 (0.0300) & 0.6169 (0.0106) & \underline{0.4689 (0.0198)}\\
    percepta & \underline{DOD}     & MLP & 0.3883 (0.0461) & 0.5599 (0.0196) & \underline{0.3781 (0.1204)}\\
    percepta & PCA$_0$ & RFR & 0.9149 (0.0015) & 0.5581 (0.0040) & 0.4691 (0.0013)\\
    percepta & L       & MLP & 0.4918 (0.0065) & 0.5493 (0.0307) & -0.1530 (0.0859)\\
    \rowcolor{revised_bg}
    percepta & BBB   & XGB & 1.0000 \text{(\;\;n.a.\;\;)} & 0.5258 \text{(\;\;n.a.\;\;)} & 0.5398 \text{(\;\;n.a.\;\;)} \\
    percepta & \underline{H} & MLP             & 0.4964 (0.0203) & 0.5069 (0.0214) & \underline{0.1712 (0.0641)}\\
    cddd     & PCA$_0$ & MLP & 0.6795 (0.0532) & 0.5036 (0.0185) & 0.5491 (0.0492)\\
    cddd     & BBB     & MLP & 0.5058 (0.1007) & 0.4734 (0.0208) & 0.2447 (0.0235)\\
    molBERT  & PCA$_0$ & MLP & 0.4517 (0.0180) & 0.3908 (0.0115) & 0.2541 (0.0458)\\
    rdkit    & BBB     & MLP & 0.5923 (0.0258) & 0.3565 (0.0373) & 0.1727 (0.0444)\\
        \hline
\end{tabular}
\end{table}

Overall, the orderings according to the validation $R^2$ values are quite comparable to what we see in the external test set, but the values themselves are almost always significantly reduced.
In light of the results shown in Table~\ref{tbl:external_test}, we can conclude that the cross-validation scheme is applicable for hyperparameter selection, but we would also like to emphasize that the use of an external test set is crucial and crossvalidation during training is not enough to characterize the generalizability of the regression models.
This latter remark is especially important in the case of datasets with a moderate number of molecules, such as this one. Therefore, our decision to exclude models with low validation results from the final evaluation was justified.

We selected the best \color{revised_text} single-task linear models and \color{black} 3 membrane specific \color{revised_text} MLP \color{black} models for further interpretability analysis.

\color{revised_text}
\section{Discussion}

\subsection{Comparision of single-tak vs. multi-task scenarios}

\begin{figure}[t]
  \caption{
  \textcolor{revised_text}
  {Comparison of the best performing (maximal R$^2$ valid) single-task and multi-task models for all tasks and input representation pairs. }
  }
  \label{fig:single_vs_multi}
  \fcolorbox{revised_bg}{revised_bg}{\includegraphics[width=0.95\linewidth]{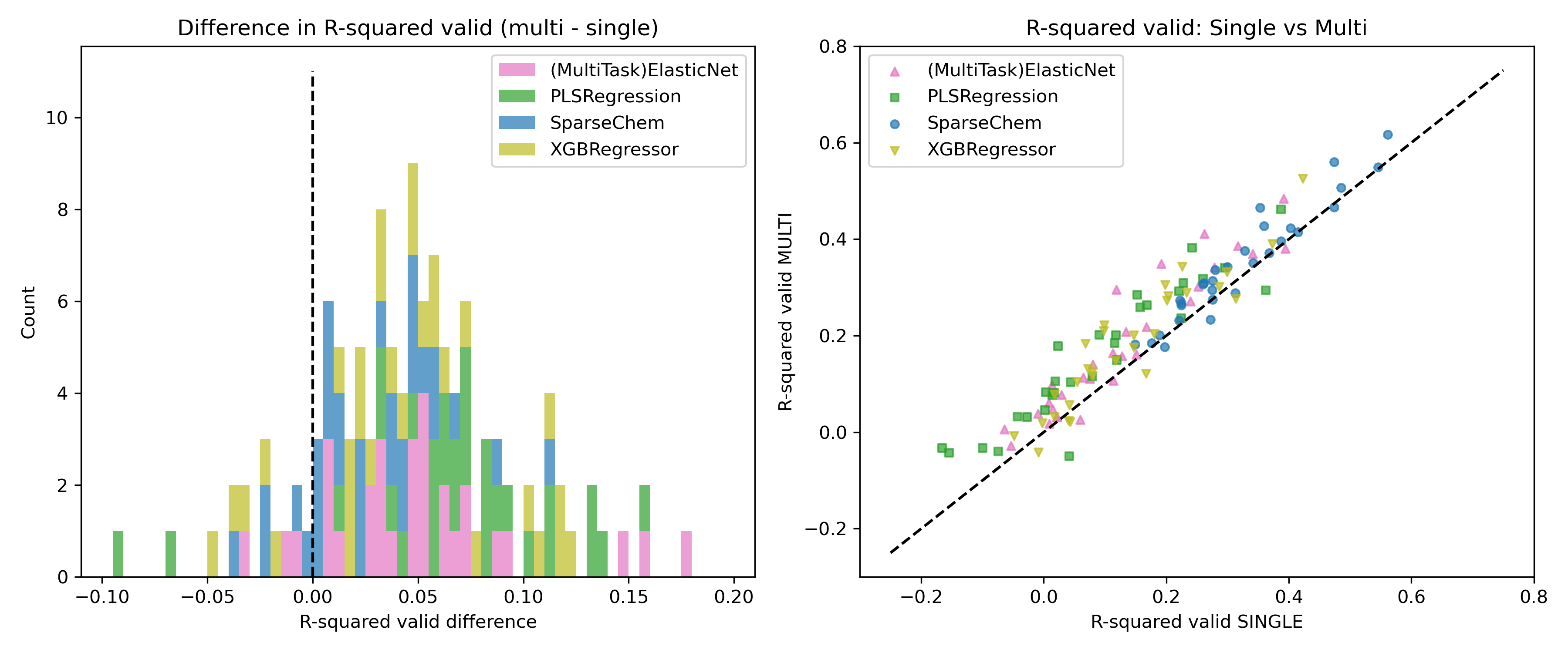   }}
\end{figure}

In case of the MLP, XGB, ElasticNet and PLSRegressor model classes we have results for both single-task and multi-task training. 
Choosing the best model for each input representation (5 possibilies) and tasks (6 possibilities), gives us $30$ tuned models both for single- and multi-task.
This results in 120 single-multi pairs given the $4$ model classes.
In Figure~\ref{fig:single_vs_multi} we can see that in 105 of the 120 cases the multi-task models outperform the corresponding single-task models.  

\subsection{Explainable feature importance}

The best single-task linear (ElasticNet) tuned models were selected to examine explainable feature importance.
A summary of performance metrics of these models can be seen in Table~\ref{tbl:linar_best} of the Supplementary material.

\begin{figure}[t]
  \caption{
  \textcolor{revised_text}
      {Feature importance (coefficients) of best single-task linear models (Elastic Net) by percepta feature for each inner validation loop of the 4-fold cross validation.
      }
  }
  \label{fig:feature_importance_elasticnet_percepta}
  \fcolorbox{revised_bg}{revised_bg}{
  \includegraphics[width=0.95\linewidth, trim={0.5cm 0cm 4.5cm 0cm},clip]{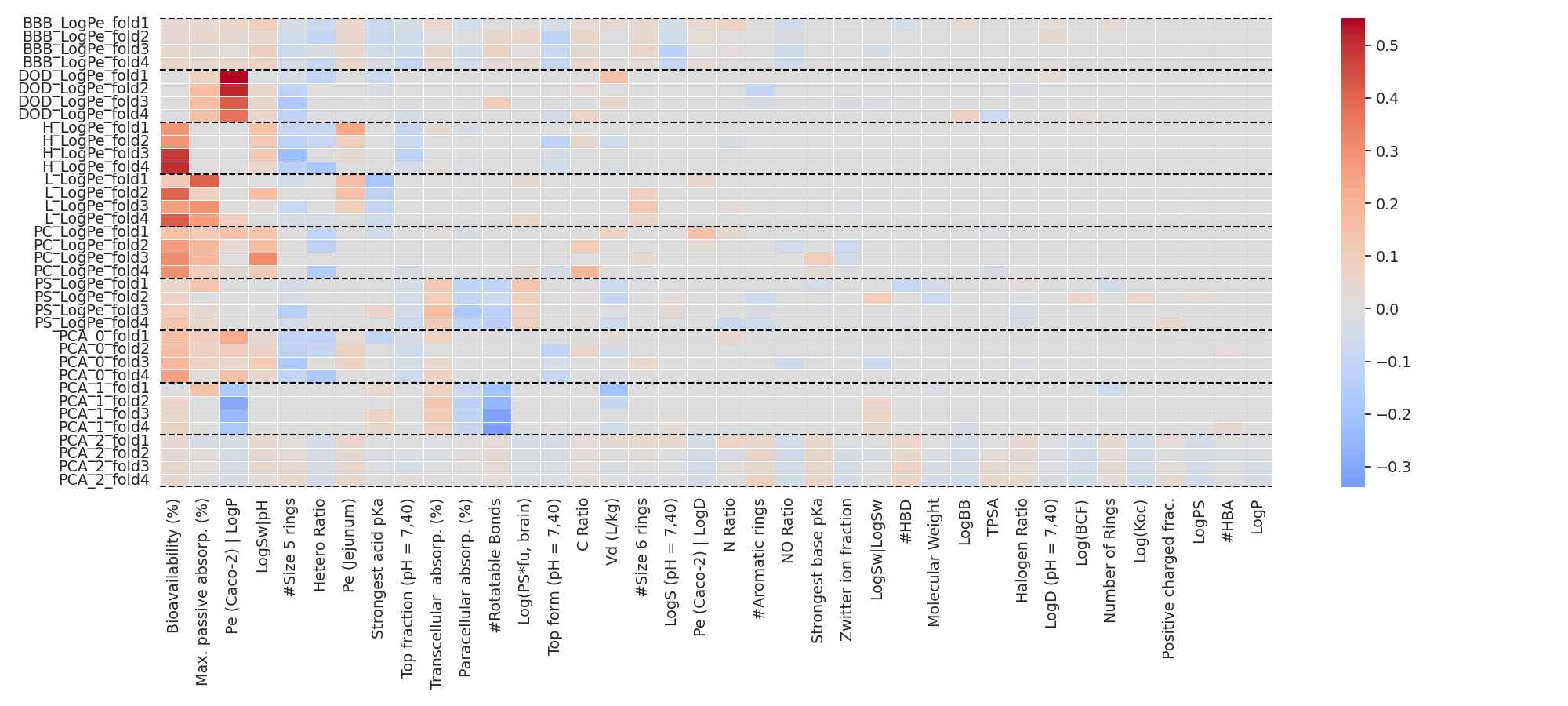}}
\end{figure}

The coefficients of the single-task linear models provide an interpretable feature importance metric over the also interpretable physchem features.
We present a heatmap of these feature importances across each target and 4-fold cross validation held out set in Figure~ \ref{fig:feature_importance_elasticnet_percepta}.
We can conclude that, with respect to the targets, the tuned models provide consistent importance vectors: the 4-row-blocks are consistent in Figure~\ref{fig:feature_importance_elasticnet_percepta}.
In general, derived features corresponding to higher level permeability predictors are selected with high weights for all tasks (e.g.: Bioavailability, Maximum passive absorbtion, Caco-2 permeability). These use physchem properties like logP, logD and pKa, and this explains why otherwise expected features are not used by our model (explaining away effect due to colinearity). 

As expected the features selected by the $PCA_0$ models also include the above mentioned higher level derived features. The later PCA components utilize noticeably different features. 

\color{black}

\subsection{Physicochemical profiles of the best-fit models}

\begin{figure}[!ht]
  \caption{Venn diagrams (left) and violin plots (middle and right) of the 10 lowest (left respectively) and the 10 highest (right respectively) penetrating compounds on heart (H)-, brain (BBB)-specific membrane and dodecane (DOD). Violin plots show the distribution of the compounds with respect to their physicochemical properties.}
  \label{fig:violin}
  \includegraphics[width=0.17\linewidth]{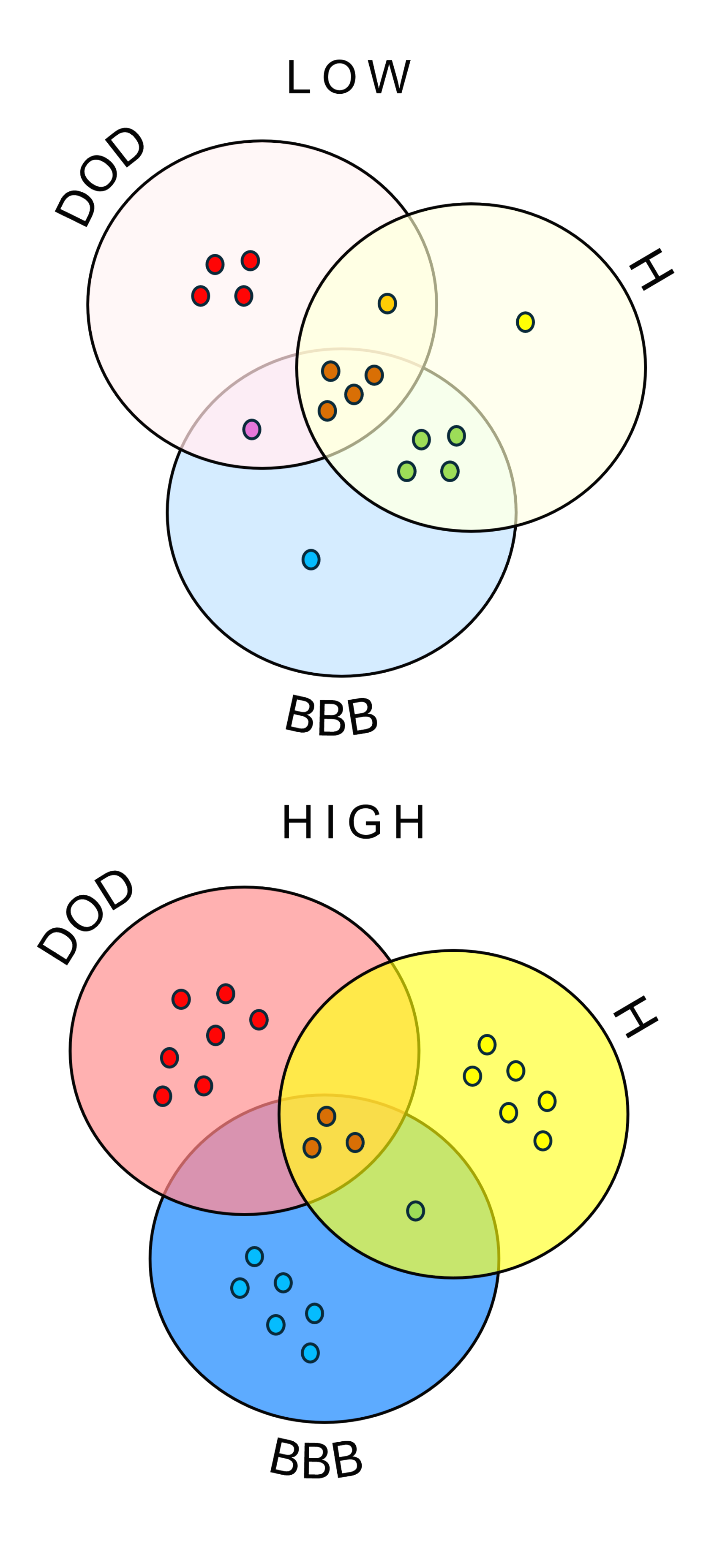}
  \includegraphics[width=0.6\linewidth]{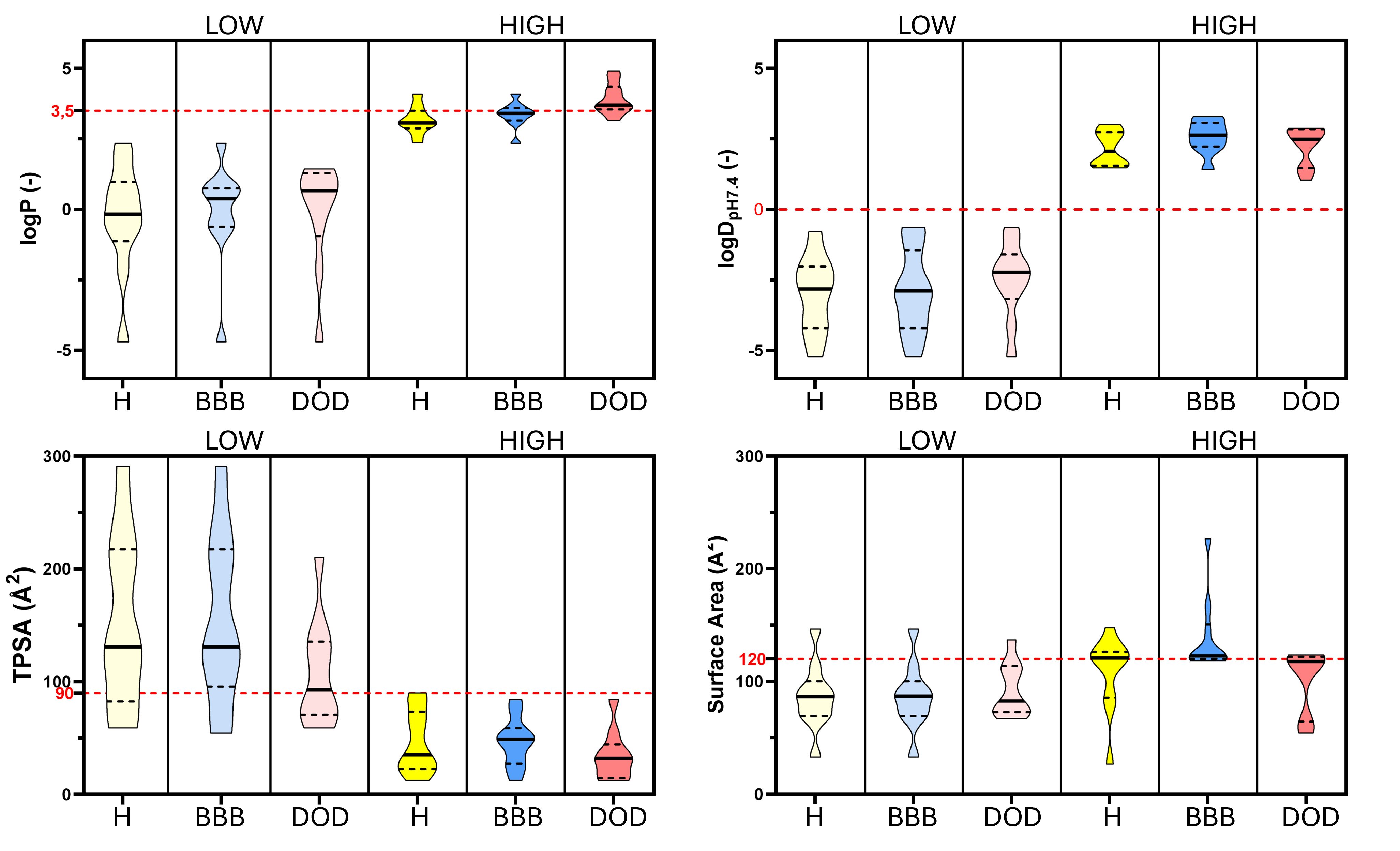}
\end{figure}

Based on Table~\ref{tbl:external_test} the overall best performing \color{revised_text} membrane specific MLP \color{black} models (H, BBB, DOD) were further analyzed by collecting the 10 best (high-$\star$) and 10 worst (low-$\star$) penetrating molecules across the corresponding membranes. 
Thus, roughly 15\% of the dataset was further investigated.

In Figure~\ref{fig:violin} the distribution of the physicochemical characteristics of these compounds are depicted using violin plots. 
It is worth noting that many low-penetrating compounds are shared between the membranes, whereas high-penetrating compounds appeared to be unique for the specific membranes (See Figure~\ref{fig:violin}, Venn diagram). 
Based on these results, we propose interpretable selectivity criteria for the top-performing compounds.

The logP value of the high-DOD compounds were higher than for the other membranes (logP > 3.5), on the other hand, no such trend was found for logD7.4 values.
Dodecane, being a hydrophobic, is incapable of proton-dissociation thus it is not surprising that the lipophilicity of the non-ionized form of the molecules is a definitive parameter for this membrane.  
The TPSA cut-off for blood-brain barrier penetration is known to be 90 \AA$^2$. 
\cite{thai2020fast, cornelissen2023explaining, van1998estimation}
In our analysis low-BBB and low-H compounds tend to fall into a higher TPSA-region.
Most of the lowest BBB-compounds (and low-Hs) followed this pattern, however, for DOD no such rule was recognizable.

The Solvent Accessible Surface Area (SASA) gives a measure of the contact area between the molecule and solvent, typically measured by simulating the contact between the van der Waals surface of the molecule and a sphere representing the van der Waals radius of the solvent.
\cite{mitternacht2016freesasa} 
It is mostly used for estimating protein-ligand interactions.
\cite{konstantinidis2021estimation, ausaf2014review} 
Generally, when a molecule has a higher SASA value, it is more exposed to the solvent (or membrane lipids), thus being more prone to participate in various molecular interactions. 
Although here the high-penetrators for all membranes presented the same median value, this value also became a sharp border: high-BBB compounds typically scattered in the ``upper region'' (SASA>120 \AA$^2$), while for dodecane they scattered in the ``lower region'' (SASA<120 \AA$^2$). 
SASA can also be helpful in predicting hydrophopic/hydrophilic balance critical for membrane partitioning.
\cite{zhu2017entropy} 
These results point towards the presence of some intramolecular interactions enabled by elevated SASA mostly governing the passive transport through the brain-specific phospholipids (containing both hydrophobic and hydrophilic moieties) in contrast to the penetration through the hydrophobic dodecane. 
Finally, high-H compounds are present both in the upper and lower SASA region.

\color{revised_text}
\section{Conclusion}
\color{black}

Our results highlight several key findings regarding the predictability of membrane permeability across various molecular representations and membranes. 
The most accurate predictions were achieved for the first principal component (PCA$_0$) of the permeability measurements, suggesting that this component captures the dominant and most learnable variance across the studied compounds and membranes.
Only carrying out experiments on multiple membranes allows us to study PCA$_0$ and conclude its higher predictability than any membrane permeability alone.
This sheds light on the value of our unique multitask dataset.

The models for PCA$_1$ and PCA$_2$ yielded positive validation $R^2$ values, suggesting the presence of predictable differences between membrane types. 
However, their low magnitudes point to the limited predictive power under the current setup.
\color{revised_text}
The feature importance values extracted from linear models further confirm the differences between assays.
\color{black}
Representations that yield high accuracy in general may still perform poorly for particular cases like liver membrane permeability. 
This further underlines the distinct characteristics of different membrane profiles and the challenge of developing truly generalizable models.
In the particular case of the PS membrane we experienced difficulties during both \textit{in vitro}, and \textit{in silico} modeling, therefore we suggest exploring other possible negatively charged PAMPA membranes.  

For this study, we decided to focus on 6 artificial membranes, and completed the full measurement matrix with the same molecules to examine separate and joint modeling. 
Although the decision to include several membranes on a fixed budget meant that the effective number of molecules measured is lower than in other similar studies, the overall size of the published dataset is, to our knowledge, one of the largest non-sparse multimembrane datasets. 
Examining the 
\textcolor{revised_text}{
    performance differences between corresponding single- and multi-task models}, we see that multi-task dominate, both in the case of MLP models and in classical ML models.

Another notable finding is that, the traditional expert-designed Percepta features almost always outperform the
high-dimensional representations derived from pre-trained models: CDDD and MolBERT.
Furthermore, since their internal structure is inherently difficult to interpret, their use offers little practical benefit in this context.
This leads us to advise caution when using high-dimensional molecular representations in low-sample-size scenarios. 
The particularly poor performance of the also high-dimensional ECFP, a widely accepted gold standard,
further supports this claim and highlights the risks of overfitting in such settings.

Furthermore, we observed that the set of highly penetrating molecules in case of different membranes show less simmilarity, then those of the lowest penetrating ones, and some phisicocamical properties correlate with the differences between the highly penetrating molecules.

Finally, we hypothesize that the gap between Percepta and RDKit features are partly attributable to RDKit’s lack of a pK$_a$ prediction model, which limits its ability to capture pH-dependent protonation states. 
\textcolor{revised_text}{
    Note that, even though the training of MolBERT involved physicocemical descriptors from RDKit as auxiliary tasks, it was never trained with pKa supervision.
}
This could be a critical shortcoming for modeling permeability, as ionization can significantly influence a compound’s membrane-crossing behavior.

\section{Author Contributions}
Conceptualization: Gy.B., A.A. 
Funding acquisition, and Resources: Gy.B., A.A, Y.M. 
Methodology, and Data interpretation: A.F., A.V., Gy.B., A.A.
Planning and carrying out of wet lab measurements: A.V., R.B.
Investigation: A.F., A.V., R.B., A.A.
Data curation, Formal analysis, Visualization, and Validation: A.F., A.V.
Software and Modeling,: A.F.
Writing - original draft: A.F., A.V., A.A.
Writing - review and editing: Y.M., Gy.B.

\begin{acknowledgement}

The authors thank would like to thank founding of the following:
(1) CELSA - Active Learning (CELSA/21/019, Projectcode: 3E210628);
(2) Leuven.AI - KU Leuven institute for AI, B-3000, Leuven, Belgium;
(3) Marie Skłodowska-Curie grant agreement No. 956832;

We thank to Balázs Volk for providing some of the compounds for the study.

\end{acknowledgement}

\section{Conflict of interest statement}
The authors declare no competing financial interests.

\newpage
\section{Data and Software Availability}
All source codes of the project are available at https://github.com/Formandras/pampa\_qsar.
Additionally, the following files are available free of charge.

Essential for reproducibility using our code:
\begin{itemize}
  \item cddd.csv: precalculated CDDD descriptors for all molecules. Full matrix of compound IDs, CV-fold of compund and one colomn for each CDDD feature. 
  \item ecfp.csv: precalculated ECFP descriptors for all molecules. Compound IDs, and CV-fold listed for all compunds. Due to the sparse nature 3rd column only contains the list of indices active input features for each molecule.
  \item molBERT.csv: precalculated MolBERT descriptors for all molecules. Full matrix of compound IDs, CV-fold of compund and one colomn for each MolBERT feature.
  \item percepta.csv: precalculated Percepta descriptors for all molecules. Full matrix of compound IDs, CV-fold, and all percepta features corresponding to the column names. 
  \item percepta\_columns.json: normalisation values (er column) for Percepta features
  \item rdkit.csv: precalculated RDKit descriptors for all molecules. Full matrix of compound IDs, CV-fold, and all RDKit features corresponding to the column names. 
  \item rdkit\_columns.json: normalisation values (er column) for RDKit features
  \item all\_plate\_measurements.csv: datamatrix containing all repeats for all molecules on all targets.
  \item all\_desalted\_smiles\_measurements\_folds.csv: precalculated merged datamatrix of all information contained in all\_plate\_measurements\_mean.csv and all\_plate\_desalted\_smiles.csv 
\end{itemize}
Descriptors generated using external software, raw measurement matrices and middle steps to prepare essential files above:
\begin{itemize}
  
  \item CELSA\_No1\_Pe\_MR.xlsx: Raw measurement data of Plate 1.
  \item CELSA\_No2\_Pe\_MR.xlsx: Raw measurement data of Plate 2.
  \item CELSA\_No3\_Pe\_MR.xlsx: Raw measurement data of Plate 3.
  \item CELSA\_No4\_Pe\_MR.xlsx: Raw measurement data of Plate 4.
  \item CELSA\_No5\_Pe\_MR.xlsx: Raw measurement data of Plate 5.
  \item CELSA\_No6\_Pe\_MR.xlsx: Raw measurement data of Plate 6.
  \item CELSA\_No7\_Pe\_MR.xlsx: Raw measurement data of Plate 7.
  \item CELSA\_No8\_Pe\_MR.xlsx: Raw measurement data of Plate 8.
  \item CELSA\_No9\_Pe\_MR.xlsx: Raw measurement data of Plate 9.
  
  \item all\_plate\_desalted\_smiles.csv: precalculated datamatrix containing wet lab experimental notes of the molecules and the desalted, canonical SMILES 
  \item all\_plate\_measurements\_mean.csv: precalculated datamatrix containing average values of repeated measurements from all\_plate\_measurements.csv
  \item all\_plate\_percepta\_raw.csv: Percepta output to generate all\_plate\_percepta.csv
  \item all\_plate\_percepta.csv: cleared view of all\_plate\_percepta\_raw.csv before transformation to training compatible percepta.csv
  
  \item CELSA\_Mcule\_BTRG\_egyesitett.xlsx: in-house dataset of drug-like molecules of the BUTE: Biomimetic Technologies Research Group
  \item "Molekulabank SE GYTK GYKI.xlsx": in-house dataset of drug-like molecules of the SU: Department of Pharmaceutical Chemistry

\end{itemize}

\section{Abbreviations}

\begin{tabular}{ l l }
 ADME & absorption, distribution, metabolism and excretion \\
 API & Active Pharmaceutical Ingredients \\
 BERT & Bidirectional Encoder Representations from Transformers \\
 BPLE & Brain Polar Lipid Extract \\
 Caco-2 & passive permeability across Caco-2 cell monolayers \\
 CDDD & Continuous and Data-Driven molecular Descriptors \\
 CNS & Central Nervous System \\
 CV & Cross-Validation \\
 BBB & blood–brain barrier \\ 
 DOD & Dodecane \\
 DTR & Decision Tree Regressor \\
 ECFP & extended-connectivity fingerprints \\  
 ECFP6 & extended-connectivity fingerprints calculated with a radius of 3 \\  
 HPLC & High-performance liquid chromatography \\
 HPLE & Heart Polar Lipid Extracts \\
 LogD & distribution coefficient of ionized species of a chemical compound between the lipid and \\
  & aqueous phases \\
 LogP & partition coefficient of an ionized species molecule between aqueous and lipophilic phases \\
 LogP$_e$ & log of effective permeability \\
 LogS & aqueous solubility of an ionized species of molecule \\
 LPLE & Liver Polar Lipid Extracts \\
 MR & Membrane Retention \\
 MLP & Multi Layer Percepton \\
 MTEN & MultiTaskElasticNet \\
 PAMPA & parallel artificial membrane permeability assay \\
\end{tabular}

\begin{tabular}{ l l } 
 PC & Phosphatidylcholine \\
 PCA & Principal Component Analysis \\ 
 PCA$_0$ & First Principal Component \\ 
 PCA$_1$ & Second Principal Component \\ 
 PCA$_2$ & Third Principal Component \\
 P$_e$ & effective permeability \\
 pK$_a$ & -log$_{10}$ of the proton dissociation constant \\
 PL & phospholipid \\
 PLD & Partial least-squares \\
 PS & Phosphatidylserine \\
 QSPR & Quantitative Structure Property Relationship \\
 $R^2$ & Coefficient of Determination \\
 RFR & Random Forest Regressor \\
 RMSE & Root Mean Squared Error \\
 SMILES & Simplified Molecular Input Line Entry System \\
 SVR & Support Vector Regression \\
 
\end{tabular}

\begin{suppinfo}

\begin{itemize}
  \item PamPaper\_supporting.pdf: PDF file containing most of the supporting figures, tables and lists.
  \item \textcolor{revised_text}{
    all\_plate\_desalted\_smiles\_folds\_with\_scaffolds.csv: the calculated scaffold for each molecule, using the `MakeScaffoldGeneric' function of the `Chem.Scaffolds.MurckoScaffold' library of RDKit.
    }
    
  \item \textcolor{revised_text}{
  applicability\_Percepta\_BBB\_MLP\_fold\_1234-combined.pdf: example  applicability domain (AD) exploration using residual plots (scatter plot of the squared error vs. input features) for BBB ElasticNet trained on Percepta features.
  }
\end{itemize}
\end{suppinfo}

\newpage
\bibliography{pampa}


\newpage

\renewcommand{\thefigure}{S\arabic{figure}} 
\renewcommand{\thetable}{S\arabic{table}}   
\setcounter{figure}{0} 
\setcounter{table}{0}  

\subsection{Modeling hyperparameters}
The modeling classes and hyperparameters (with the grid values in [square brackets]) were as follows:
 
\textbf{DecisionTreeRegressor}: decision tree regressor (DTR) 
    \begin{itemize}
    \item \textit{min\_samples\_leaf}: The minimum number of samples required to be at a leaf node
    [~1,~2,~4,~8,~16,~32,~64~]
    \item \textit{min\_samples\_split}: minimum number of samples required to split an internal node
    [~2,~4,~8,~16,~32,~64~]
    \end{itemize}
    
\textbf{RandomForestRegressor}: random forest regressor (RFR) ensemble of 1,000 DTR models 
    
    \begin{itemize}
    \item \textit{min\_samples\_leaf}: same as \textit{DecisionTreeRegressor}
    \item \textit{min\_samples\_split}: same as \textit{DecisionTreeRegressor}
    \item \textit{max\_features}: the proportion of features to consider when looking for the best split
    [~0.1,~0.2,~0.4,~0.8,~1.0~]
    \end{itemize}

\textbf{ElasticNet}: linear regression implementation with $L_1$ (Lasso) and $L_2$ (Ridge) type regularization.
    \begin{itemize}
    \item \textit{alpha}: parameter that scales the total regularization term 
    [
        0, 0.01, 0.05, 0.1, 0.5, 1, 5, 10, 50, 100, 500, 1000
    ]
    \item \textit{l1\_ratio}: the ratio of Lasso to the total $(\frac{L_1}{total})$
    [
        0.0, 0.1, 0.2, 0.3, 0.4, 0.5, 0.6, 0.7, 0.8, 0.9, 1.0 
    ]
    \end{itemize}
    
\textbf{MultiTaskElasticNet}: joint optimization of the ElasticNet to multiple targets 

    \begin{itemize}
    \item \textit{alpha}: same as \textit{ElasticNet}
    \item \textit{l1\_ratio}: same as \textit{ElasticNet}
    \end{itemize}
    
\textbf{BayesianRidge}: fits a Bayesian Linear Regression with Ridge ($L_2$ type) regularization

\textbf{PLSRegression}: Partial least-squares regression \cite{PLS1986} commonly used \cite{vincze2021corneal} cross-decomposition algorithm in chemoinformatics
    
    \begin{itemize}
    \item \textit{n\_components}: number of components
    [1, 2, 5, 10, 20, 50, 100]

    \end{itemize}
    
\textbf{SVR}: Epsilon-Support Vector Regression model
    
    \begin{itemize}
    \item \textit{kernel}: kernel type  
    [
        `linear', 
        `rbf',
        `sigmoid'
        `poly'
    ]
    \item \textit{gamma}: kernel coefficient
    [
        `scale',
        `auto'
    ]
    \item \textit{C}: regularization parameter
    [
        0.001,
        0.01,
        0.1,
        1,
        10
    ]
    \item \textit{epsilon}: specifies the epsilon-tube
    [
        0.0001,
        0.001,
        0.01,
        0.1,
        1,
        10,
        100
    ]
    \end{itemize}
For the polynomial kernel `poly', degrees of 1, 2, and 3 were considered.

\color{revised_text}
\textbf{XGBoost}: XGBoost is an optimized distributed gradient boosting library designed to be highly efficient, flexible and portable. It implements machine learning algorithms under the Gradient Boosting framework. \cite{chen2016xgboost,xgboost-python}
    
    \begin{itemize}

    \item \textit{objective}: regression with squared loss
    ["reg:squarederror"],
    \item \textit{n\_estimators}: number of trees
    [100, 1000],
    \item \textit{max\_depth}: Maximum depth of a tree.
    [4, 5, 6],
    \item \textit{lambda}: L2 regularization term on weights
    [0.001, 0.01, 0.1, 1, 10],
    \item \textit{alpha}: L1 regularization term on weights
    [0.001, 0.01, 0.1, 1, 10]
    \item \textit{subsample}: Subsample ratio of the training instances. Setting it to 0.5 means that XGBoost would randomly sample half of the training data prior to growing trees. and this will prevent overfitting. Subsampling will occur once in every boosting iteration.
    [0.5, 1.0]

    \end{itemize}
\color{black}

\textbf{MLP}: the SparseChem\cite{sparsechem2022} model is a Multi-Layer Perceptron (\textbf{MLP}) implementation specialized for sparse input and output matrices
    \begin{itemize}
    \item \textit{number of layers} [1, 2]
    \item \textit{number of neurons} in each layer [20, 50, 100]
    \item \textit{dropout} ratio [0.6, 0.8]
    \item \textit{weight decay} ($L_2$ type regularization) [0.01, 0.1]
    \item \textit{learning rate} [0.1, 0.3]
    \end{itemize}

\newpage
\subsection{Model selection for testing}
\begin{figure}[ht]
  \caption{
    Figures based on Figure~\ref{fig:all_model_best}, Visual explanation of which \textit{tuned models} are further discussed in Tables~\ref{tbl:absolute_best},~\ref{tbl:sklearn_best},~\ref{tbl:PCA_BBB_features}, and~\ref{tbl:sklearn_PCA}, respectively in this order from top to bottom. 
    Circled \textit{tuned models} were further selected for test-set evaluation, squared ones were not.}
  \label{fig:chose_model_visual}
  \includegraphics[width=0.8\linewidth]{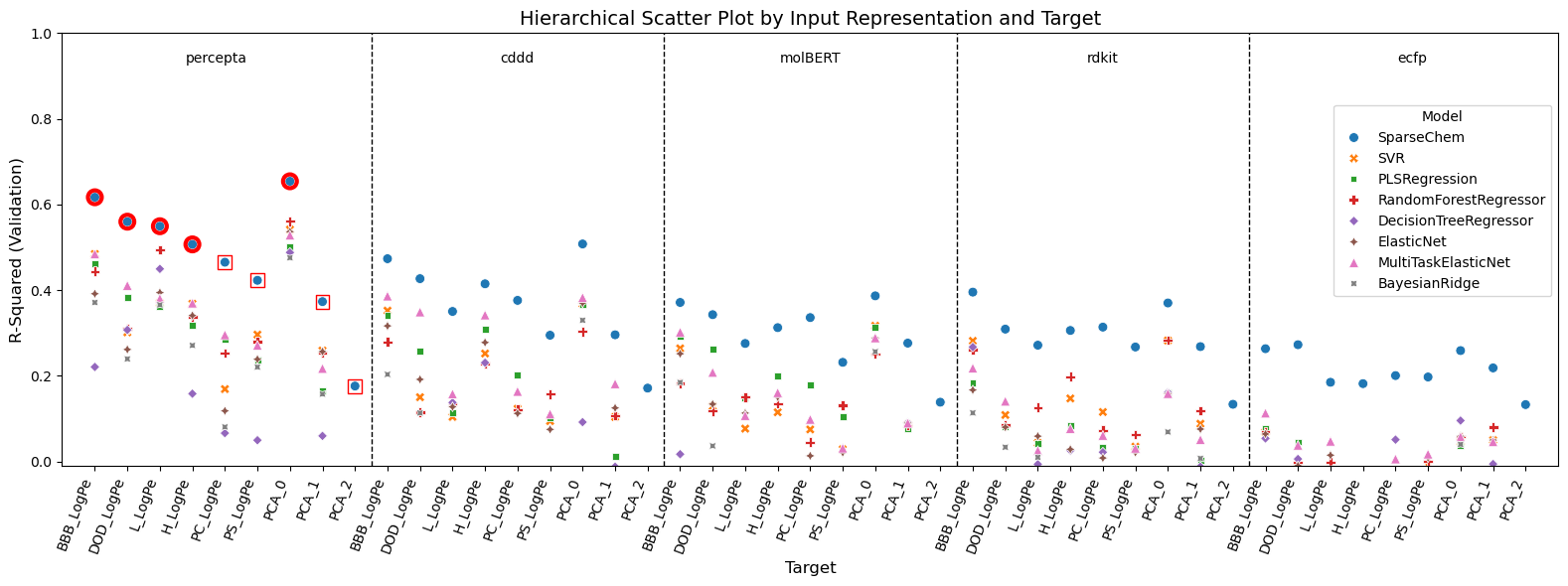}
  \includegraphics[width=0.8\linewidth]{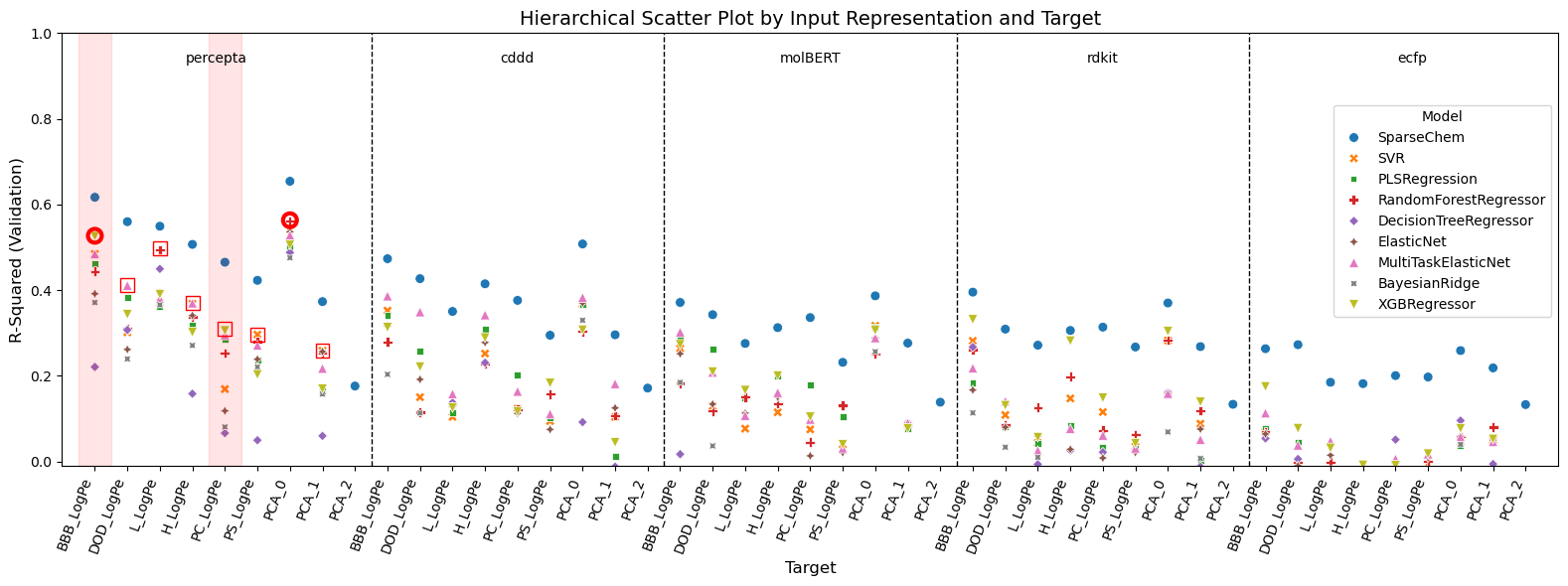}
  \includegraphics[width=0.8\linewidth]{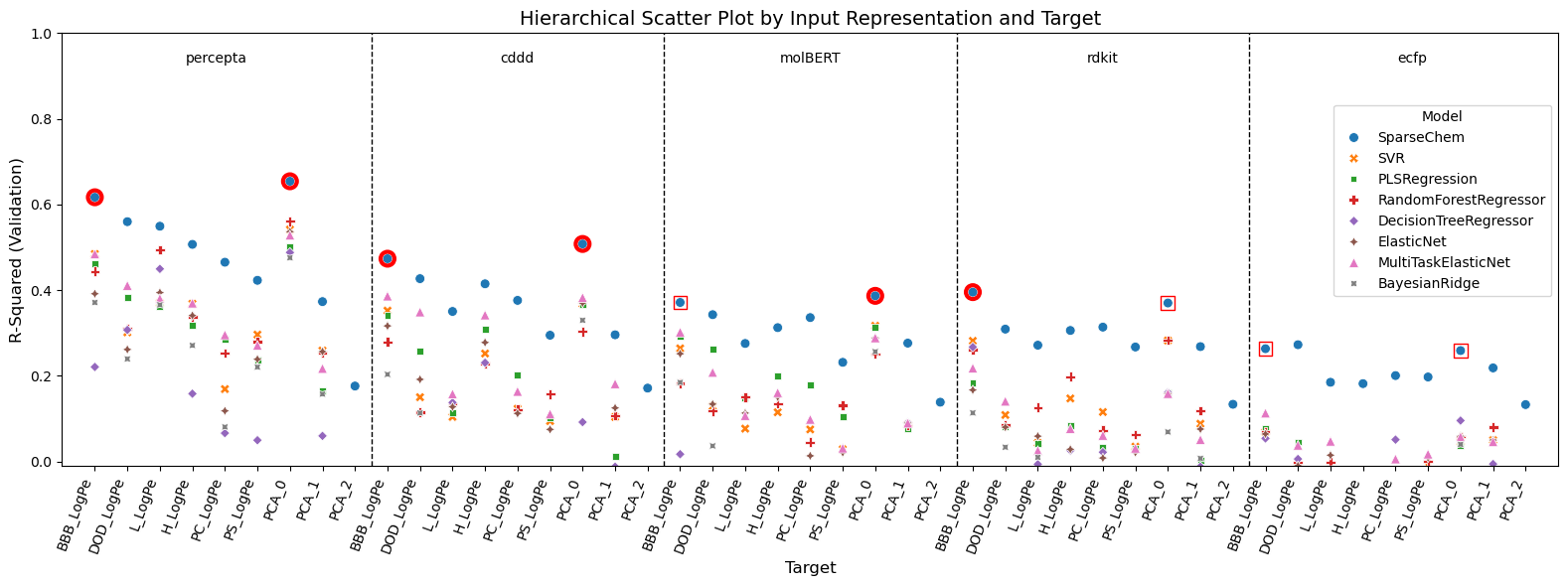}
  \includegraphics[width=0.8\linewidth]{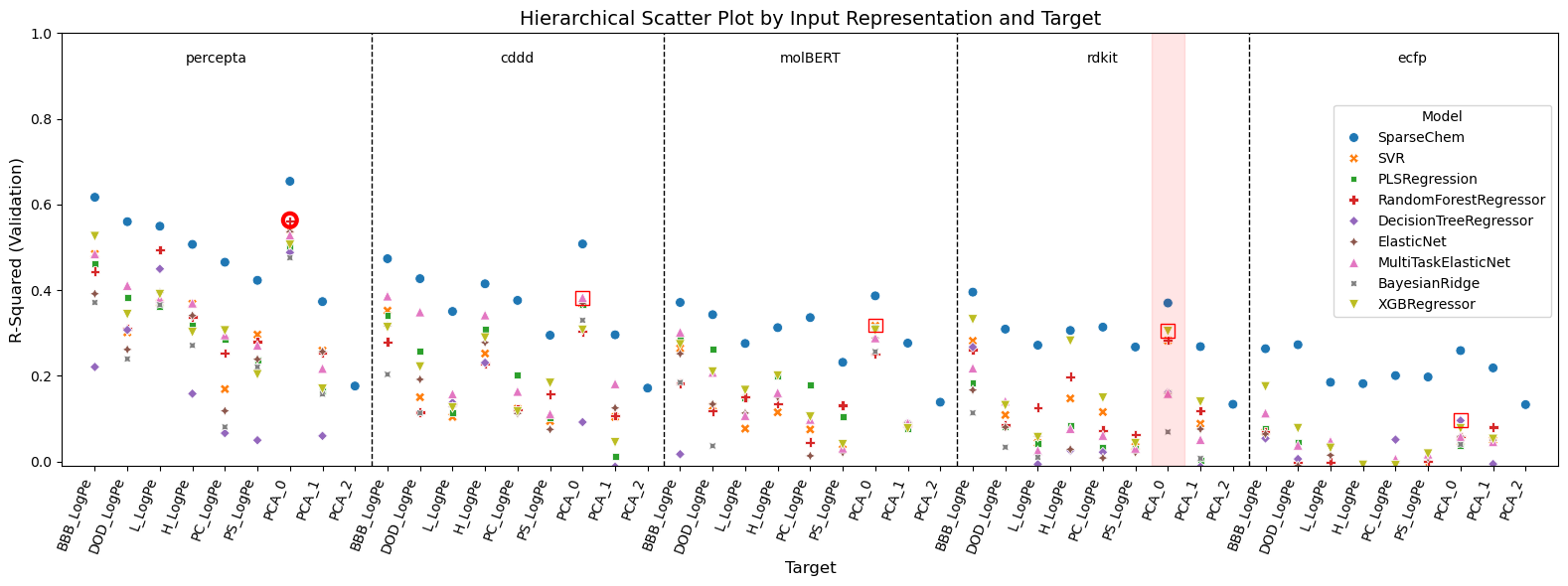}
\end{figure}

\newpage
\subsection{Compound library and Cross Validation}
\label{sup:comp_sel}

\begin{figure}[t]
  \caption{4-fold Cross Validation compound sets and randomly selected hold-out test set.}
  \label{fig:CV_split}
  \includegraphics[width=0.75\linewidth]{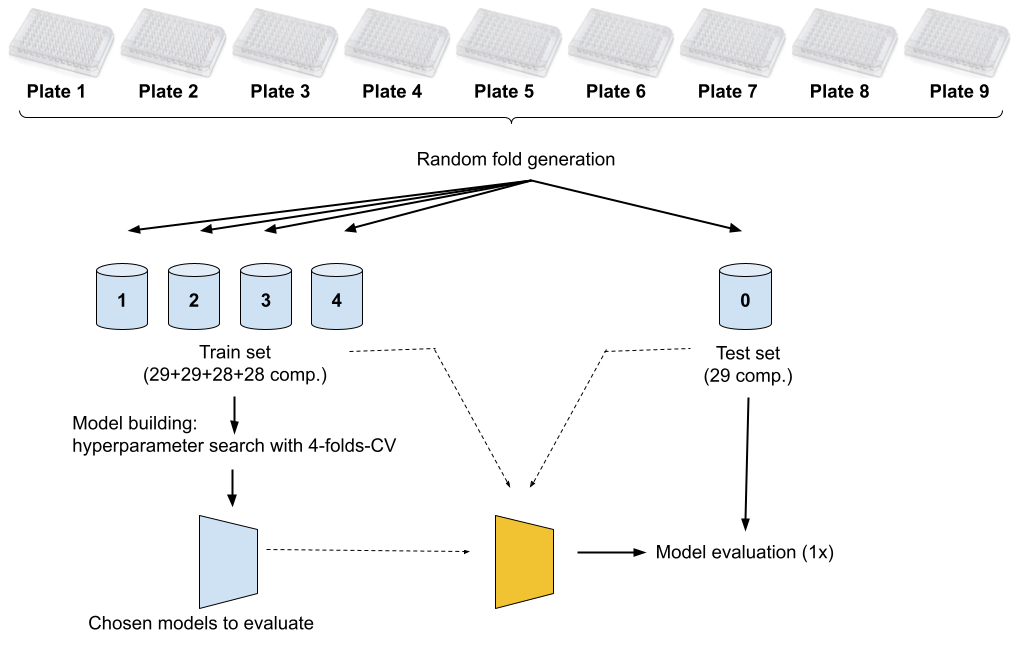}
\end{figure}

\label{sup:desalt}
Salt removal and standardization was done using RDKit v'2023.09.6' in the following way:
\begin{algorithm}
\caption{Desalting}\label{alg:desalt}
\begin{algorithmic}
\State $mol \gets  rdkit.Chem.MolStandardize.rdMolStandardize.Cleanup(mol)$
\State $desalted \gets rdkit.Chem.SaltRemover.SaltRemover(defnData=salts).StripMol(mol)$
\State $frags \gets rdkit.Chem.GetMolFrags(desalted, asMols=True)$

\If{$len(frags) > 1$} 
    \State $desalted \gets chooseMostSignifficantFragment(frags)$
\EndIf 

\State $desalted \gets rdkit.Chem.RemoveStereochemistry(desalted)$
\State $desalted \gets rdkit.Chem.MolStandardize.rdMolStandardize.Uncharger().uncharger$ $.uncharge(desalted)$
\State $desalted \gets rdkit.Chem.MolStandardize.rdMolStandardize.Cleanup(desalted)$
\end{algorithmic}
\end{algorithm}

\newpage

\subsection{Hyperparameters of tuned models}
\begin{table}[ht]
  \caption{ Hyperparameters corresponding to Table~\ref{tbl:absolute_best}.  }
  \label{sup:HP_absolute_best}
    \begin{tabular}{llll}
    \hline
    eval target & input & training target & parameters \\
    \hline
    L\_LogPe & percepta &  \shortstack{'L\_LogPe', 'PS\_LogPe', \\'DOD\_LogPe', 'PC\_LogPe', \\'H\_LogPe', 'BBB\_LogPe'} & \shortstack{'hidden\_sizes': [100], \\'weight\_decay': 0.01, \\'dropouts\_trunk': [0.8], \\'lr': 0.1} \\
    \hline
    PS\_LogPe & percepta & \shortstack{'L\_LogPe', 'PS\_LogPe', \\'DOD\_LogPe', 'PC\_LogPe', \\'H\_LogPe', 'BBB\_LogPe'} & \shortstack{'hidden\_sizes': [100], \\'weight\_decay': 0.01, \\'dropouts\_trunk': [0.5], \\'lr': 0.3} \\
    \hline
    DOD\_LogPe & percepta & \shortstack{'L\_LogPe', 'PS\_LogPe', \\'DOD\_LogPe', 'PC\_LogPe', \\'H\_LogPe', 'BBB\_LogPe'} & \shortstack{'hidden\_sizes': [50, 50], \\'weight\_decay': 0.1, \\'dropouts\_trunk': [0.6, 0.6], \\'lr': 0.3} \\
    \hline
    PC\_LogPe & percepta & \shortstack{'L\_LogPe', 'PS\_LogPe', \\'DOD\_LogPe', 'PC\_LogPe', \\'H\_LogPe', 'BBB\_LogPe'} & \shortstack{'hidden\_sizes': [20], \\'weight\_decay': 0.01, \\'dropouts\_trunk': [0.6], \\'lr': 0.3} \\
    \hline
    H\_LogPe & percepta & \shortstack{'L\_LogPe', 'PS\_LogPe', \\'DOD\_LogPe', 'PC\_LogPe', \\'H\_LogPe', 'BBB\_LogPe'} & \shortstack{'hidden\_sizes': [50], \\'weight\_decay': 0.01, \\'dropouts\_trunk': [0.5], \\'lr': 0.3} \\
    \hline
    BBB\_LogPe & percepta & \shortstack{'L\_LogPe', 'PS\_LogPe', \\'DOD\_LogPe', 'PC\_LogPe', \\'H\_LogPe', 'BBB\_LogPe'} & \shortstack{'hidden\_sizes': [20], \\'weight\_decay': 0.01, \\'dropouts\_trunk': [0.6], \\'lr': 0.3} \\
    \hline
    PCA\_0 & percepta & \shortstack{PCA\_0 of 'L\_LogPe', 'PS\_LogPe', \\'DOD\_LogPe', 'PC\_LogPe', \\'H\_LogPe', 'BBB\_LogPe'} & \shortstack{'hidden\_sizes': [50, 50], \\'weight\_decay': 0.1, \\'dropouts\_trunk': [0.5, 0.5], \\'lr': 0.3} \\
    \hline
    PCA\_1 & percepta & \shortstack{PCA\_1 of 'L\_LogPe', 'PS\_LogPe', \\'DOD\_LogPe', 'PC\_LogPe', \\'H\_LogPe', 'BBB\_LogPe'} & \shortstack{'hidden\_sizes': [50], \\'weight\_decay': 0.01, \\'dropouts\_trunk': [0.5], \\'lr': 0.3} \\
    \hline
    PCA\_2 & percepta & \shortstack{PCA\_2 of 'L\_LogPe', 'PS\_LogPe', \\'DOD\_LogPe', 'PC\_LogPe', \\'H\_LogPe', 'BBB\_LogPe'} & \shortstack{'hidden\_sizes': [50], \\'weight\_decay': 0.01, \\'dropouts\_trunk': [0.6], \\'lr': 0.1} \\
    \hline
    \end{tabular}
\end{table}

\begin{table}
  \caption{ Hyperparameters corresponding to Table~\ref{tbl:sklearn_best}. }
  \label{sup:HP_sklearn_best}
    \begin{tabular}{lllll}
    \hline
    eval target & model & input & training target & parameters \\
    \hline
    L & RFR & percepta & 'L\_LogPe' & \shortstack{'n\_estimators': 1000, \\'min\_samples\_split': 32, \\'max\_features': 0.8} \\
    \hline
    PS & SVR & percepta & 'PS\_LogPe' & \shortstack{'kernel': 'rbf', \\'gamma': 'scale', \\'C': 1, \\'epsilon': 0.0001} \\
    \hline
    DOD & MTEN & percepta & \shortstack{'L\_LogPe', 'PS\_LogPe', \\'DOD\_LogPe', 'PC\_LogPe', \\'H\_LogPe', 'BBB\_LogPe'} & \shortstack{'alpha': 0.5, \\'l1\_ratio': 0.2} \\
    \hline
    \rowcolor{revised_bg}
    PC & XGB & percepta & \shortstack{'L\_LogPe', 'PS\_LogPe', \\'DOD\_LogPe', 'PC\_LogPe', \\'H\_LogPe', 'BBB\_LogPe'} & \shortstack{'objective': 'reg:squarederror', \\'n\_estimators': 1000, \\'subsample': 0.5, \\'max\_depth': 4, \\'lambda': 0.1, 'alpha': 10}	 \\
    \hline
    H & MTEN & percepta & \shortstack{'L\_LogPe', 'PS\_LogPe', \\'DOD\_LogPe', 'PC\_LogPe', \\'H\_LogPe', 'BBB\_LogPe'} & \shortstack{'alpha': 0.1, \\'l1\_ratio': 1.0} \\
    \hline
    \rowcolor{revised_bg}
    BBB & XGB & percepta & 'BBB\_LogPe' & \shortstack{'objective': 'reg:squarederror', \\'n\_estimators': 100, \\'subsample': 0.5, \\'max\_depth': 6,\\ 'lambda': 0.1, 'alpha': 0.01} \\
    \hline
    PCA\_0 & RFR & percepta & \shortstack{PCA\_0 of 'L\_LogPe', 'PS\_LogPe', \\'DOD\_LogPe', 'PC\_LogPe', \\'H\_LogPe', 'BBB\_LogPe'} & \shortstack{'n\_estimators': 1000, \\'min\_samples\_split': 4, \\'max\_features': 0.1} \\
    \hline
    PCA\_1 & SVR & percepta & \shortstack{PCA\_1 of 'L\_LogPe', 'PS\_LogPe', \\'DOD\_LogPe', 'PC\_LogPe', \\'H\_LogPe', 'BBB\_LogPe'} & \shortstack{'kernel': 'rbf', \\'gamma': 'scale', \\'C': 1, \\'epsilon': 0.1} \\
    \hline
    PCA\_2 & SVR & percepta & \shortstack{PCA\_2 of 'L\_LogPe', 'PS\_LogPe', \\'DOD\_LogPe', 'PC\_LogPe', \\'H\_LogPe', 'BBB\_LogPe'} & \shortstack{'kernel': 'poly', \\'degree': 1, \\'gamma': 'scale', \\'C': 1, \\'epsilon': 0.01} \\
    \hline
    \end{tabular}
\end{table}

\begin{table}
  \caption{ Hyperparameters corresponding to Table 
  \ref{tbl:PCA_BBB_features}. }
  \label{sup:HP_PCA_BBB_features}
    \begin{tabular}{lllllr}
    \hline
    eval target & model & input & training target & parameters \\\hline
    PCA\_0 & MLP & percepta & \shortstack{PCA\_0 of ['L\_LogPe', \\'PS\_LogPe', 'DOD\_LogPe', \\'PC\_LogPe', 'H\_LogPe', \\'BBB\_LogPe']} & \shortstack{'hidden\_sizes': [50, 50], \\'weight\_decay': 0.1, \\'dropouts\_trunk': [0.5, 0.5], \\'lr': 0.3} \\ \hline
    PCA\_0 & MLP & cddd & \shortstack{PCA\_0 of ['L\_LogPe', \\'PS\_LogPe', 'DOD\_LogPe', \\'PC\_LogPe', 'H\_LogPe', \\'BBB\_LogPe']} & \shortstack{'hidden\_sizes': [15], \\'weight\_decay': 0.01, \\'dropouts\_trunk': [0.5], \\'lr': 0.1} \\ \hline
    PCA\_0 & MLP & molBERT & \shortstack{PCA\_0 of ['L\_LogPe', \\'PS\_LogPe', 'DOD\_LogPe', \\'PC\_LogPe', 'H\_LogPe', \\'BBB\_LogPe']} & \shortstack{'hidden\_sizes': [100], \\'weight\_decay': 0.01, \\'dropouts\_trunk': [0.6], \\'lr': 0.1} \\ \hline
    PCA\_0 & MLP & rdkit & \shortstack{PCA\_0 of ['L\_LogPe', \\'PS\_LogPe', 'DOD\_LogPe', \\'PC\_LogPe', 'H\_LogPe', \\'BBB\_LogPe']} & \shortstack{'hidden\_sizes': [50], \\'weight\_decay': 0.1, \\'dropouts\_trunk': [0.6], \\'lr': 0.3} \\ \hline
    PCA\_0 & MLP & ecfp & \shortstack{PCA\_0 of ['L\_LogPe', \\'PS\_LogPe', 'DOD\_LogPe', \\'PC\_LogPe', 'H\_LogPe', \\'BBB\_LogPe']} & \shortstack{'hidden\_sizes': [50], \\'weight\_decay': 0.01, \\'dropouts\_trunk': [0.6], \\'lr': 0.3} \\ \hline

    BBB\_LogPe & MLP & percepta  & \shortstack{'L\_LogPe', 'PS\_LogPe', \\'DOD\_LogPe', 'PC\_LogPe', \\'H\_LogPe', 'BBB\_LogPe'} & \shortstack{'hidden\_sizes': [20], \\'weight\_decay': 0.01, \\'dropouts\_trunk': [0.6], \\'lr': 0.3} \\ \hline
    BBB\_LogPe & MLP & cddd      & \shortstack{'BBB\_LogPe'}                                                                      & \shortstack{'hidden\_sizes': [15], \\'weight\_decay': 0.01, \\'dropouts\_trunk': [0.5], \\'lr': 0.3} \\ \hline
    BBB\_LogPe & MLP & rdkit     & \shortstack{'L\_LogPe', 'PS\_LogPe', \\'DOD\_LogPe', 'PC\_LogPe', \\'H\_LogPe', 'BBB\_LogPe'} & \shortstack{'hidden\_sizes': [20], \\'weight\_decay': 0.01, \\'dropouts\_trunk': [0.6], \\'lr': 0.3} \\ \hline
    BBB\_LogPe & MLP & molBERT   & \shortstack{'L\_LogPe', 'PS\_LogPe', \\'DOD\_LogPe', 'PC\_LogPe', \\'H\_LogPe', 'BBB\_LogPe'} & \shortstack{'hidden\_sizes': [100], \\'weight\_decay': 0.1, \\'dropouts\_trunk': [0.6], \\'lr': 0.1} \\ \hline
    BBB\_LogPe & MLP & ecfp      & \shortstack{'L\_LogPe', 'PS\_LogPe', \\'DOD\_LogPe', 'PC\_LogPe', \\'H\_LogPe', 'BBB\_LogPe'} & \shortstack{'hidden\_sizes': [50], \\'weight\_decay': 0.01, \\'dropouts\_trunk': [0.6], \\'lr': 0.3} \\ \hline
    \end{tabular}
\end{table}

\begin{table}
  \caption{ Hyperparameters corresponding to Table~\ref{tbl:sklearn_PCA}. }
  \label{sup:HP_sklearn_PCA}
    \begin{tabular}{lllllr}
    \hline
    eval target & model & input & training target & parameters \\\hline
    PCA\_0 & RFR    & percepta  & \shortstack{PCA\_0 of ['L\_LogPe', \\'PS\_LogPe', 'DOD\_LogPe', \\'PC\_LogPe', 'H\_LogPe', \\'BBB\_LogPe']}                     & \shortstack{'n\_estimators': 1000, \\'min\_samples\_split': 4, \\'max\_features': 0.1} \\ \hline
    PCA\_0 & RFR    & percepta  & \shortstack{PCA\_0 of ['L\_LogPe', \\'PS\_LogPe', 'DOD\_LogPe', \\'PC\_LogPe', 'H\_LogPe', \\'BBB\_LogPe']}                     & \shortstack{'n\_estimators': 1000, \\'min\_samples\_split': 4, \\'max\_features': 0.1} \\ \hline
    PCA\_0 & MTEN   & cddd      & \shortstack{  ['PCA\_0', 'PCA\_1', 'PCA\_2']  \\ of ['L\_LogPe', \\'PS\_LogPe', 'DOD\_LogPe', \\'PC\_LogPe', 'H\_LogPe', \\'BBB\_LogPe']} & \shortstack{'alpha': 0.5, \\'l1\_ratio': 0.2} \\ \hline
    PCA\_0 & SVR    & molBERT   & \shortstack{PCA\_0 of ['L\_LogPe', \\'PS\_LogPe', 'DOD\_LogPe', \\'PC\_LogPe', 'H\_LogPe', \\'BBB\_LogPe']}                     & \shortstack{'kernel': 'linear', \\'C': 0.01, \\'epsilon': 0.0001} \\ \hline
    \rowcolor{revised_bg}
    PCA\_0 & XGB    & rdkit     & \shortstack{PCA\_0 of ['L\_LogPe', \\'PS\_LogPe', 'DOD\_LogPe', \\'PC\_LogPe', 'H\_LogPe', \\'BBB\_LogPe']}                     & \shortstack{'objective': 'reg:squarederror', \\'n\_estimators': 100, \\'max\_depth': 5,\\ 'lambda': 0.1, 'alpha': 1}	 \\ \hline
    PCA\_0 & DTR    & ecfp      & \shortstack{PCA\_0 of ['L\_LogPe', \\'PS\_LogPe', 'DOD\_LogPe', \\'PC\_LogPe', 'H\_LogPe', \\'BBB\_LogPe']}                     & \shortstack{'min\_samples\_split': 64} \\ \hline
    \end{tabular}

\end{table}

\begin{table}
  \caption{
  Performance of the best models evaluated on the test set. 
  Find corresponding R$^2$ values in Table~\ref{tbl:external_test}.
  }
  \label{tbl:external_test_corr}
\begin{tabular}{llrrrl}
        \hline
    input & target & corr train & corr valid & corr test\\
        \hline
    percepta & PCA$_0$   & 0.7732 (0.0250) & 0.8101 (0.0101) & 0.7139 (0.0147) & MLP\\
    percepta & BBB & 0.8062 (0.0171) & 0.7979 (0.0073) & 0.6929 (0.0140) & MLP\\
    percepta & DOD & 0.6716 (0.0132) & 0.7636 (0.0168) & 0.6858 (0.0623) & MLP\\
    percepta & PCA$_0$   & 0.9691 (0.0008) & 0.7610 (0.0026) & 0.7047 (0.0009) & RFR\\
    percepta & L   & 0.7281 (0.0111) & 0.7525 (0.0251) & 0.1937 (0.0378) & MLP\\
    \rowcolor{revised_bg}
    percepta & BBB   & 1.0000 \text{(\;\;n.a.\;\;)} & 0.7417 \text{(\;\;n.a.\;\;)} & 0.7359 \text{(\;\;n.a.\;\;)} & XGB\\
    percepta & H   & 0.7266 (0.0138) & 0.7167 (0.0211) & 0.4650 (0.0406) & MLP\\
    cddd     & PCA$_0$     & 0.8446 (0.0299) & 0.7314 (0.0109) & 0.7932 (0.0461) & MLP\\    
    cddd     & BBB & 0.7205 (0.0801) & 0.7043 (0.0169) & 0.5229 (0.0390) & MLP\\
    molBERT  & PCA$_0$     & 0.7169 (0.0169) & 0.6596 (0.0120) & 0.5719 (0.0444) & MLP\\
    rdkit    & BBB & 0.8024 (0.0206) & 0.6208 (0.0305) & 0.4506 (0.0485) & MLP\\
        \hline
\end{tabular}
\end{table}

\color{revised_text}
\subsection{Scaffold landscape}
The provided file 
`all\_plate\_desalted\_smiles\_folds\_with\_scaffolds.csv'
contains the calculated scaffold for each molecules, using the `MakeScaffoldGeneric' function of the `Chem.Scaffolds. 
MurckoScaffold' library of RDKit.

\begin{table}
  \caption{
    \textcolor{revised_text}{
          Repeating scaffolds
    }
  }
  \label{tbl:scaffold_repeats}
\begin{tabular}{llrrrl}
        \hline
    \rowcolor{revised_bg}
        scaffold\_smiles & fold & orig\_stock & compound\_name & plate\_number \\
        \hline
    \rowcolor{revised_bg}
        \mytextsize{8}{CC1CCC2C(C)CCC(C)(C)C2C1C} & 4 & BTRG & PGY0072 & 4 \\
    \rowcolor{revised_bg}
        \mytextsize{8}{CC1CCC2C(C)CCC(C)(C)C2C1C} & 4 & BTRG & PGY0216\_1 & 6 \\ 
    \rowcolor{revised_bg}
        \mytextsize{8}{CCC(C)C1(C)CCC2C3CCC4CC(C)CCC4(C)C3C(C)CC21C} & 3 & BTRG & Hydrocortisone & 6 \\
    \rowcolor{revised_bg}
        \mytextsize{8}{CCC(C)C1(C)CCC2C3CCC4CC(C)CCC4(C)C3C(C)CC21C} & 4 & BTRG & Prednisolone & 1 \\
    \rowcolor{revised_bg}
        \mytextsize{8}{CCC(C)C1(C)CCC2C3CCC4CC(C)CCC4(C)C3C(C)CC21C} & 4 & BTRG & Prednisone & 6 \\
    \rowcolor{revised_bg}
        \mytextsize{8}{CCC1CC(C(C)C)C(C)C2CC(C)C(C3CCCCC3)CC12} & 0 & BTRG & Enoxacin & 1 \\
    \rowcolor{revised_bg}
        \mytextsize{8}{CCC1CC(C(C)C)C(C)C2CC(C)C(C3CCCCC3)CC12} & 0 & BTRG & Norfloxacin & 1 \\
    \rowcolor{revised_bg}
        \hline
\end{tabular}
\end{table}

\subsection{Linear Coefficient percepta descriptors full names}
For the sake of space the full names of percepta descriptors in Figure~\ref{fig:feature_importance_elasticnet_percepta} were shortened. 
Here we attach the full names of all descriptrs from left to right.
\begin{enumerate} 
    \item Bioavailability (\%) (Dose, mg = 10,00)
    \item Pe (Caco-2) with LogP, 10\^-6 cm/s (pH = 7,40, rpm = 300,00)
    \item Maximum passive absorption (\%)
    \item LogSw|pH
    \item Number of Rings (size 5)
    \item No. of Rotatable Bonds
    \item Hetero Ratio
    \item Contribution of transcellular route to absorption (\%)
    \item Pe (Jejunum), 10\^-4 cm/s
    \item 1st strongest acid pKa
    \item Contribution of paracellular route to absorption (\%)
    \item Vd (L/kg)
    \item Most common form (pH = 7,40)|Fraction
    \item Most common form (pH = 7,40)|-
    \item C Ratio
    \item Log(PS*fu, brain)
    \item Number of Rings (size 6)
    \item NO Ratio
    \item Number of Aromatic Rings
    \item LogS (pH = 7,40)
    \item Pe (Caco-2) with LogD, 10\^-6 cm/s (pH = 7,40, rpm = 300,00)
    \item LogSw|LogSw
    \item N Ratio
    \item 1st strongest base pKa
    \item No. of Hydrogen Bond Donors
    \item LogBB
    \item Log(BCF)
    \item Fraction of form +1-1 (pH = 7,40)
    \item Halogen Ratio
    \item Log(Koc)
    \item Number of Rings
    \item TPSA
    \item Molecular Weight
    \item LogPS
    \item LogD (pH = 7,40)
    \item Most common form (pH = 7,40)|+
    \item LogP
    \item No. of Hydrogen Bond Acceptors
\end{enumerate} 

\color{black}

\subsection{Interpretable feature importance studies}

\begin{table}[]
  \caption{
  \textcolor{revised_text}{
    Best single-task linear regression (ElasticNet) \textit{tuned models} on the validation set for each target after hyperparameter search with 4 folds cross-validation.
    This Table supplements Figure \ref{fig:feature_importance_elasticnet_percepta}
    }
  }
  \label{tbl:linar_best}
  \begin{tabular}{lrrrrl}
\hline
\rowcolor{revised_bg}
target & corr train & corr valid & $r^2$ train & $r^2$ valid & parameters \\
\hline
\rowcolor{revised_bg}
$BBB$ & 0.750527 & 0.674643 & 0.542300 & 0.391731 & {'alpha': 0.5, 'l1\_ratio': 0.1} \\
\rowcolor{revised_bg}
$DOD$ & 0.692909 & 0.624653 & 0.464124 & 0.261704 & {'alpha': 0.1, 'l1\_ratio': 1.0} \\
\rowcolor{revised_bg}
$H$ & 0.715914 & 0.636709 & 0.496039 & 0.340970 & {'alpha': 0.1, 'l1\_ratio': 0.8} \\
\rowcolor{revised_bg}
$L$ & 0.737502 & 0.681822 & 0.530107 & 0.394466 & {'alpha': 0.1, 'l1\_ratio': 0.8} \\
\rowcolor{revised_bg}
$PC$ & 0.551864 & 0.455126 & 0.262933 & 0.118399 & {'alpha': 0.5, 'l1\_ratio': 0.3} \\
\rowcolor{revised_bg}
$PS$ & 0.660146 & 0.516586 & 0.405330 & 0.238899 & {'alpha': 0.5, 'l1\_ratio': 0.1} \\
\rowcolor{revised_bg}
$PCA_0$ & 0.833645 & 0.758294 & 0.687807 & 0.536525 & {'alpha': 0.1, 'l1\_ratio': 1.0} \\
\rowcolor{revised_bg}
$PCA_1$ & 0.630956 & 0.527209 & 0.353954 & 0.256748 & {'alpha': 0.1, 'l1\_ratio': 0.8} \\
\rowcolor{revised_bg}
$PCA_2$ & 0.580977 & 0.309568 & 0.301764 & -0.109284 & {'alpha': 0.5, 'l1\_ratio': 0.0} \\
\hline
\end{tabular}
\end{table}

\begin{figure}[t]
    \textcolor{revised_text}{
      \caption{
      Best validation R$^2$ of linear single-task (ElasticNet) models on each input-target pair.
      Hyperparameter seatrch was done in 4-fold cross-validation. 
      This Figure supplements Figure \ref{fig:feature_importance_elasticnet_percepta}.}
  \label{sup:single_task}
      }
  \fcolorbox{revised_bg}{revised_bg}{\includegraphics[width=0.95\linewidth]{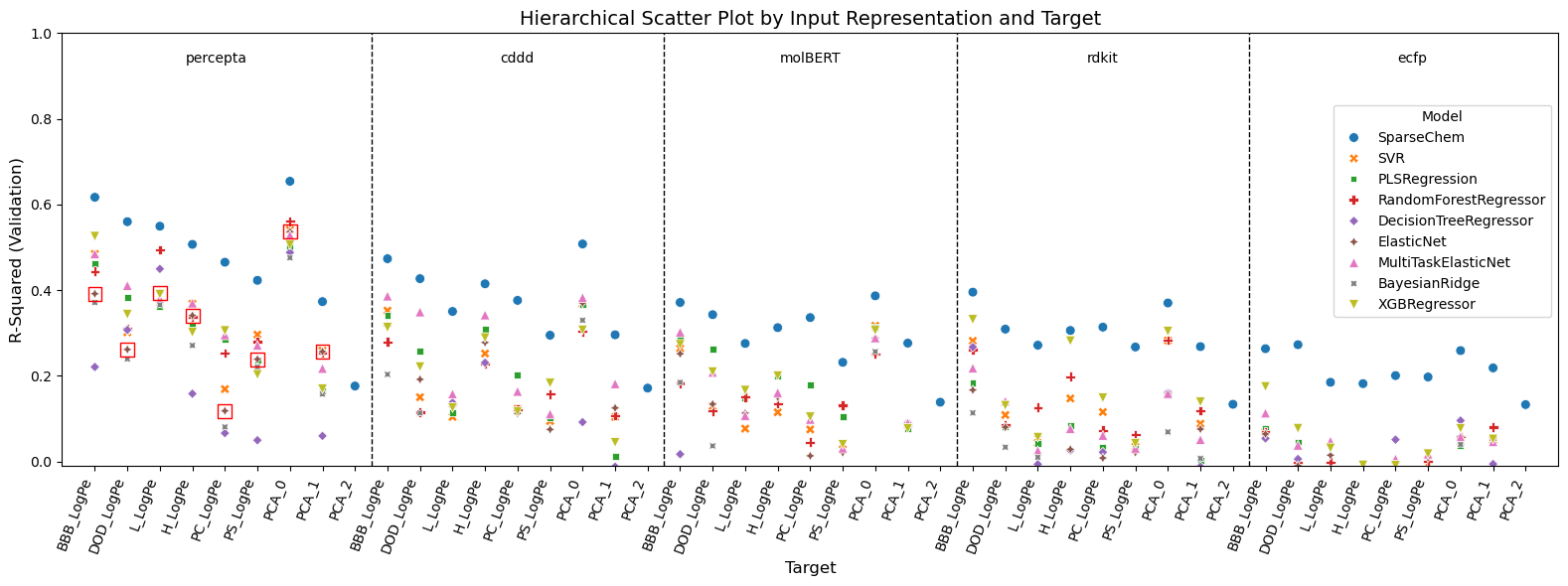}}
\end{figure}

For Figure~\ref{fig:all_model_best} style visualization of these linear models, for easier comparasion of overal performance to all tuned models, see Figure~\ref{sup:single_task}.

\newpage

\subsubsection{Additional physicochemical profile violin plots}

\begin{figure}[!ht]
  \caption{Additional violin plots (middle and right) of the 10 lowest (left respectively) and the 10 highest (right respectively) penetrating compounds on heart (H)-, brain (BBB)-specific membrane and dodecane (DOD). Violin plots show the distribution of the compounds with respect to their physicochemical properties.}
  \label{fig:violin_supp}
  \includegraphics[width=0.6\linewidth]{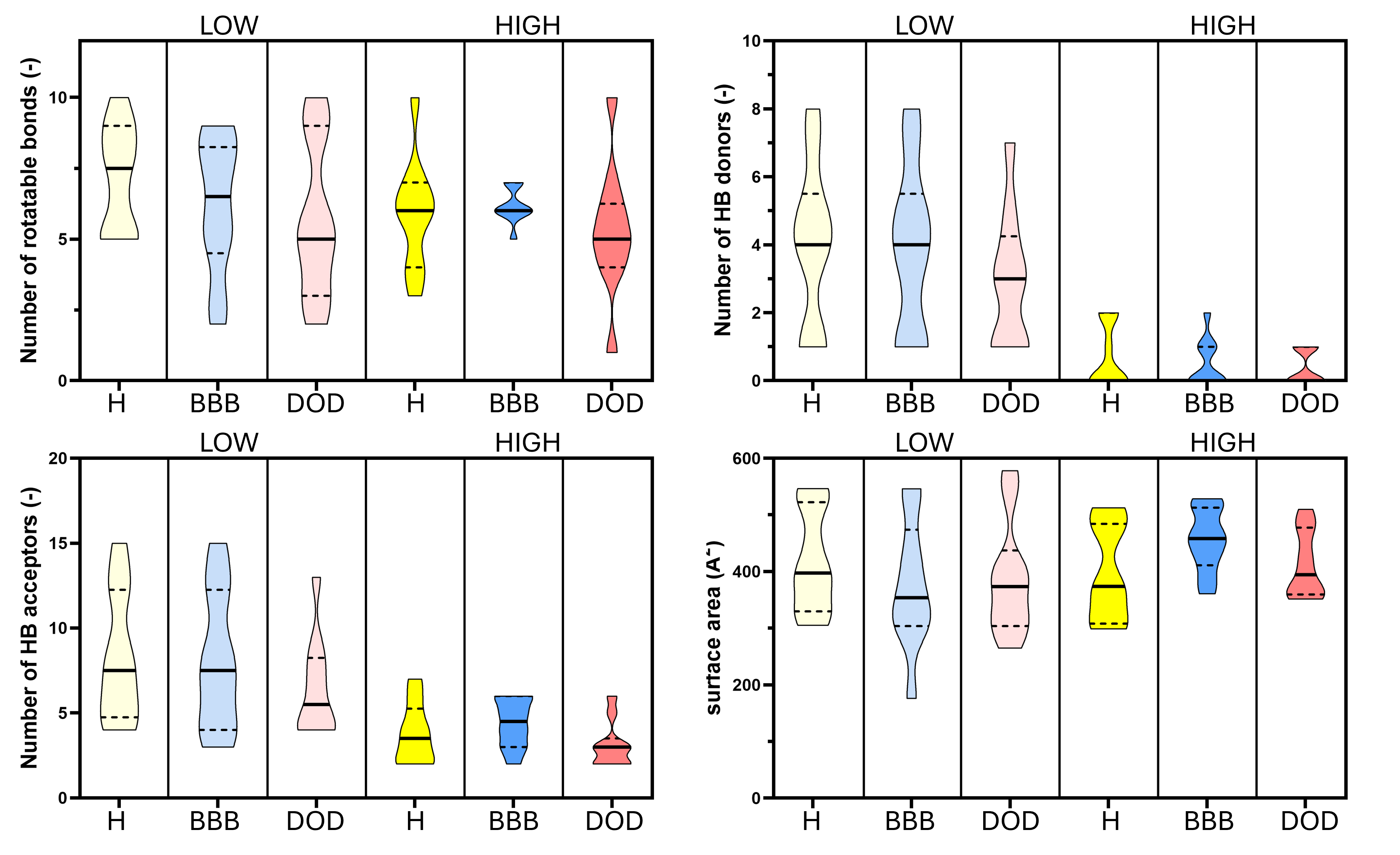}
\end{figure}

\subsection{Descriptor generation}

\label{sup:salts}
Smiles of salts removed:

"Na", "Ca", "Cl", "Br", "O", "Zn", "K", "I", "F", "N", "Li", "Mg", "O=S(=O)(O)O", "O=C(O)C(=O)O", "CS(=O)(=O)O", "O=P(=O)OO", "O=P(O)(O)O",  \\
"O=C(O)C=CC(=O)O", "O=C(O)CC(O)(CC(=O)O)C(=O)O", \\
"O=C(O)C(O)C(O)C(=O)O", "CC(C)(C)N", "NC(N)=NCCCC(N)C(=O)O", "CCO", \\
"O=S(=O)(O)CCO", "NC(CO)(CO)CO", "O=S(=O)(O)c1ccccc1", "CC(=O)O", \\
"Cc1ccc(S(=O)(=O)O)cc1", "O=C(O)O", "O=C(O)CCC(=O)O", "O=C(O)c1ccccc1", \\
"O=CO", "O=C(O)CC(O)C(=O)O", "CCN(CC)CC", "CCNCC", "CC(C)C(=O)O", \\
"NC1CCCCC1", "CCC(=O)O"

\label{sup:percepta_descriptor_names}
All 58 Percepta descriptors: "1st strongest acid pKa", "1st strongest base pKa" are derived from dropped featurs. All other features that are bald, were dropped based on lack of variance in the values.

"LogPS", 
"LogBB", 
"Log(PS*fu, brain)", 
"Pe (Jejunum), $10^{-4}$ cm/s", 
"Maximum passive absorption (\%)", 
"Contribution of transcellular route to absorption (\%)", 
"Contribution of paracellular route to absorption (\%)", 
"Molecular Weight", 
"No. of Hydrogen Bond Donors", 
"No. of Hydrogen Bond Acceptors", 
"TPSA", 
"No. of Rotatable Bonds", 
"C Ratio", 
"N Ratio", 
"NO Ratio", 
"Hetero Ratio", 
"Halogen Ratio", 
"Number of Rings", 
"Number of Aromatic Rings", 
\textbf{
"Number of Rings (size 3)", 
"Number of Rings (size 4)"}, 
"Number of Rings (size 5)", 
"Number of Rings (size 6)",
\textbf{ 
"Dielectric Constant", 
"Index of Refraction", 
"Surface Tension", 
"Density", 
"Polarizability", 
"Molar Volume", 
"Molar Refractivity"}, 
"LogS (pH = 7,40)", 
"LogSw|LogSw", 
"LogSw|pH", 
\textbf{
"BP (Pressure (mmHg) = 760,00)", 
"VP (Temperature (C) = 37,00)", 
"Enthalpy of Vaporization", 
"Flash Point"}, 
"Pe (Caco-2) with LogP, $10^{-6}$ cm/s (pH = 7,40, rpm = 300,00)", 
"Pe (Caco-2) with LogD, $10^{-6}$ cm/s (pH = 7,40, rpm = 300,00)", 
"Log(BCF)", 
"Log(Koc)", 
\textbf{
"Parachor"}, 
"LogP", 
"LogD (pH = 7,40)", 
\textbf{
"pKa(Acid)|pKa", 
"pKa(Acid)|Conf. limits", 
"pKa(Acid)|AtomNo", 
"pKa(Base)|pKa", 
"pKa(Base)|Conf. limits", 
"pKa(Base)|AtomNo"}, 
"Vd (L/kg)", 
"Fraction of form +1-1 (pH = 7,40)", 
"Most common form (pH = 7,40)|+", 
"Most common form (pH = 7,40)|-", 
"Most common form (pH = 7,40)|Fraction", 
\textbf{"
Isoelectric point"}, 
"Bioavailability (\%) (Dose, mg = 10,00)"

Columns containing Nan and columns with very small variety values were removed.
The pKa colums were converted into 2 values representing the minimal acid and the maximal base values in the correspontding lists.List of  38 Percepta descriptors after filtering and preprocessing:

"LogPS",
"LogBB",
"Log(PS*fu, brain)",
"Pe (Jejunum), $10^{-4}$ cm/s",
"Maximum passive absorption (\%)",
"Contribution of transcellular route to absorption (\%)",
"Contribution of paracellular route to absorption (\%)",
"Molecular Weight",
"No. of Hydrogen Bond Donors",
"No. of Hydrogen Bond Acceptors",
"TPSA",
"No. of Rotatable Bonds",
"C Ratio",
"N Ratio",
"NO Ratio",
"Hetero Ratio",
"Halogen Ratio",
"Number of Rings",
"Number of Aromatic Rings",
"Number of Rings (size 5)",
"Number of Rings (size 6)",
"LogS (pH = 7,40)",
"LogSw|LogSw",
"LogSw|pH",
"Pe (Caco-2) with LogP, $10^{-6}$ cm/s (pH = 7,40, rpm = 300,00)",
"Pe (Caco-2) with LogD, $10^{-6}$ cm/s (pH = 7,40, rpm = 300,00)",
"Log(BCF)",
"Log(Koc)",
"LogP",
"LogD (pH = 7,40)",
"Vd (L/kg)",
"Fraction of form +1-1 (pH = 7,40)",
"Most common form (pH = 7,40)|+",
"Most common form (pH = 7,40)|-",
"Most common form (pH = 7,40)|Fraction",
"Bioavailability (\%) (Dose, mg = 10,00)",
\textbf{
"1st strongest acid pKa",
"1st strongest base pKa"
}

All 96 RDKit Descriptors:

"rdMolDescriptors":
\footnote{https://datagrok.ai/help/domains/chem/descriptors}
[
       "CalcChi0n", "CalcChi0v", "CalcChi1n", "CalcChi1v", "CalcChi2n", "CalcChi2v", "CalcChi3n", "CalcChi3v", "CalcChi4n", "CalcChi4v",
       "CalcExactMolWt", "CalcNumAtoms", "CalcFractionCSP3", "CalcHallKierAlpha",
       "CalcKappa1", "CalcKappa2", "CalcKappa3",
       "CalcNumAliphaticCarbocycles", "CalcNumAliphaticHeterocycles", "CalcNumAliphaticRings",  "CalcNumAmideBonds", "CalcNumAromaticCarbocycles", "CalcNumAromaticHeterocycles", "CalcNumAromaticRings",
       "CalcNumHBA", \\
       "CalcNumHBD", "CalcNumHeavyAtoms", "CalcNumHeteroatoms", "CalcNumHeterocycles", "CalcNumLipinskiHBA", "CalcNumLipinskiHBD", 
       "CalcNumRings", "CalcNumRotatableBonds", "CalcNumSaturatedCarbocycles", "CalcNumSaturatedHeterocycles", "CalcNumSaturatedRings", "CalcNumSpiroAtoms", "CalcNumBridgeheadAtoms",
       "CalcPhi", "CalcTPSA", "CalcLabuteASA","BCUT2D", "CalcCrippenDescriptors", 
        ],
   
   "Descriptors": 
\footnote{https://www.rdkit.org/docs/source/rdkit.Chem.Descriptors.html}
[
       "ExactMolWt", "HeavyAtomMolWt", "MolWt", "NumValenceElectrons",
       "FpDensityMorgan1", "FpDensityMorgan2", "FpDensityMorgan3", 
       "MaxAbsPartialCharge", "MaxPartialCharge", "MinAbsPartialCharge", "MinPartialCharge",
       ],
   
   "MolSurf":  
\footnote{https://www.rdkit.org/docs/source/rdkit.Chem.MolSurf.html}
[ 
       "PEOE\_VSA1", "PEOE\_VSA2", "PEOE\_VSA3", "PEOE\_VSA4", "PEOE\_VSA5", "PEOE\_VSA6", "PEOE\_VSA7", "PEOE\_VSA8", "PEOE\_VSA9", \\
       "PEOE\_VSA10", "PEOE\_VSA11", "PEOE\_VSA12", "PEOE\_VSA13", \\
       "PEOE\_VSA14", 
       "SMR\_VSA1", "SMR\_VSA2", "SMR\_VSA3", "SMR\_VSA4", \\
       "SMR\_VSA5", "SMR\_VSA6", "SMR\_VSA7", "SMR\_VSA8", "SMR\_VSA9", \\
       "SMR\_VSA10",  
       "SlogP\_VSA1", "SlogP\_VSA2", "SlogP\_VSA3", "SlogP\_VSA4", \\
       "SlogP\_VSA5", "SlogP\_VSA6", "SlogP\_VSA7", "SlogP\_VSA8", "SlogP\_VSA9", \\
       "SlogP\_VSA10", "SlogP\_VSA11", "SlogP\_VSA12",
       "LabuteASA", "TPSA", \\
       "pyLabuteASA"]

\end{document}